%% file: main.tex
\documentclass[10pt,twocolumn,letterpaper]{article}

\usepackage{cvpr}              

\input{preamble}

\usepackage{graphicx}
\usepackage{amsmath}
\usepackage{amssymb}
\usepackage{booktabs}
\usepackage{amsfonts}  
\usepackage{amssymb}
\usepackage{xspace}
\usepackage{threeparttable}
\usepackage{multicol}
\usepackage{multirow}
\usepackage{colortbl}
\usepackage{scrextend} 

\deffootnote{2em}{1.6em}{\thefootnotemark\hspace{1em}} 

\definecolor{cvprblue}{rgb}{0.21,0.49,0.74}
\usepackage[pagebackref,breaklinks,colorlinks,citecolor=cvprblue]{hyperref}

\newcommand{\myparagraph}[1]{{\vspace{.2em} \noindent \bf #1}}
\definecolor{lightgray}{gray}{.9}
\definecolor{lightred}{RGB}{255,182,193} 
\definecolor{lightgreen}{RGB}{144,238,144} 
\definecolor{lightblue}{RGB}{173,216,230} 
\definecolor{firstcolor}{RGB}{255,188,188}
\definecolor{secondcolor}{RGB}{181, 234, 234}

\newcommand*{\affaddr}[1]{#1} 
\newcommand*{\affmark}[1][*]{\textsuperscript{#1}}

\usepackage{float}

\usepackage[capitalize]{cleveref}
\crefname{section}{Sec.}{Secs.}
\Crefname{section}{Section}{Sections}
\Crefname{table}{Table}{Tables}
\crefname{table}{Tab.}{Tabs.}

\def\paperID{****} 
\def\confName{CVPR}
\def\confYear{2024}

\begin{document}

\title{CCEdit: Creative and Controllable Video Editing via Diffusion Models}



\author{Ruoyu Feng\affmark[1,2 *], Wenming Weng\affmark[1,2], Yanhui Wang\affmark[1,2],\\  
Yuhui Yuan\affmark[2], Jianmin Bao\affmark[2],
Chong Luo\affmark[2 \dag], Zhibo Chen\affmark[1 \dag], Baining Guo\affmark[2]\\
\affaddr{\affmark[1]University of Science and Technology of China} 
\affaddr{\affmark[2]Microsoft Research Asia} 
\\ 
\affaddr{\url{https://ruoyufeng.github.io/CCEdit.github.io/}} 
}

\twocolumn[{
\renewcommand\twocolumn[1][]{#1}
\maketitle

\begin{center}
    \vspace{-10pt}
    \includegraphics[width=0.93\linewidth]{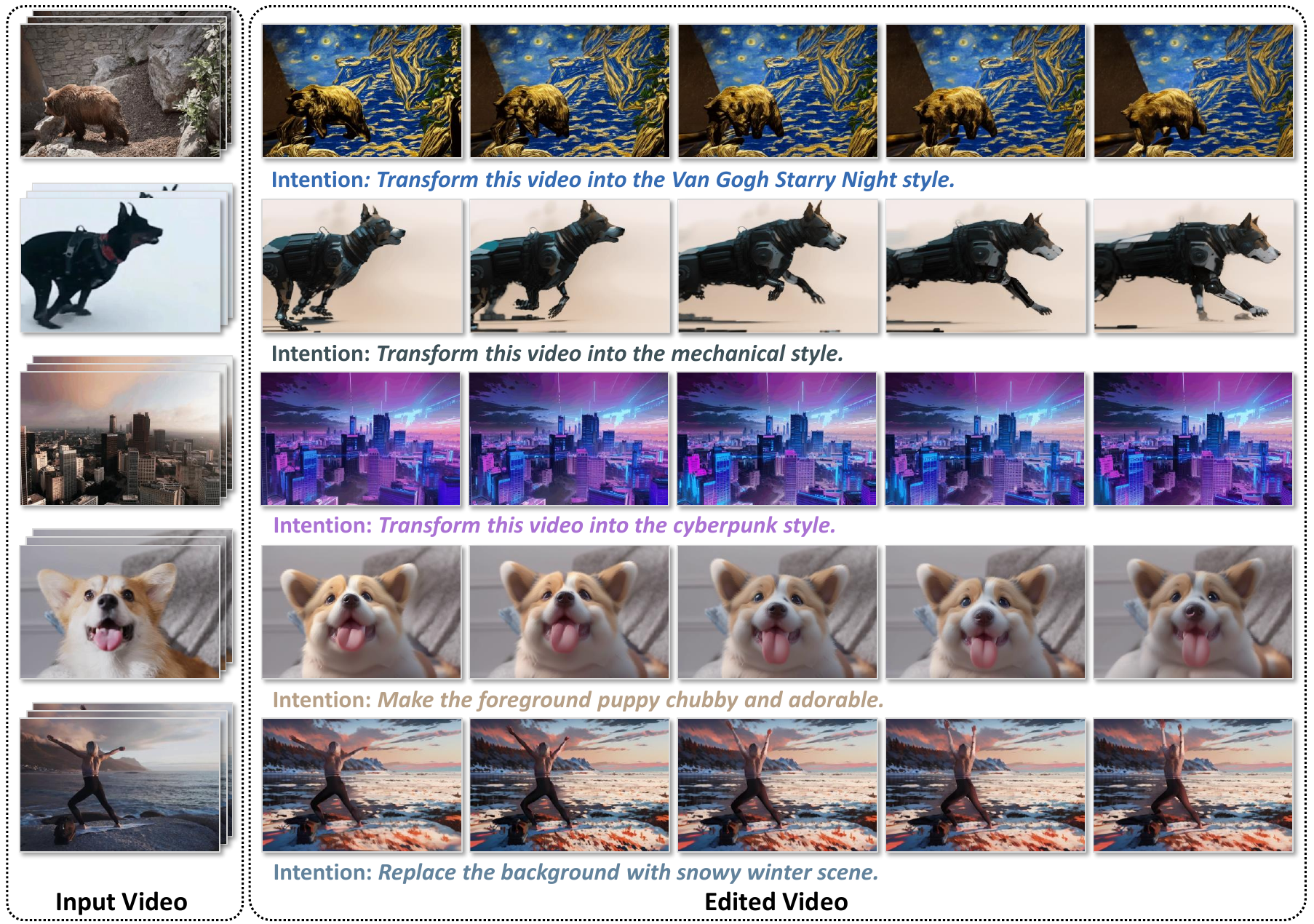}
    \vspace{-4pt}
    \captionsetup{type=figure}
    \caption{
    Built upon diffusion models, CCEdit provides users with a powerful and flexible set of video editing capabilities, including style transfer (row 1-3), foreground modifications (row 4), and background replacement (row 5).
    }
    \label{fig:teaser}
\end{center}
}]

\let\thefootnote\relax\footnotetext{$^*$~This work is done when Ruoyu Feng is an intern with MSRA.\\
 $^{\dag}$~Corresponding author.}
 
\input{sec/0_abstract}    
\input{sec/1_intro}
\input{sec/2_related_work}
\input{sec/3_approach}
\input{sec/4_benchmark}
\input{sec/5_experiments}
\input{sec/6_limitation}
\input{sec/7_conclusion}
{
    \small
    \bibliographystyle{ieeenat_fullname}
    \bibliography{main}
}
\input{sec/X_suppl}


\end{document}

%% file: preamble.tex
\usepackage[dvipsnames]{xcolor}


%% file: sec/0_abstract.tex
\begin{abstract}
In this paper, we present CCEdit, a versatile generative video editing framework based on diffusion models. Our approach employs a novel trident network structure that separates structure and appearance control, ensuring precise and creative editing capabilities. Utilizing the foundational ControlNet architecture, we maintain the structural integrity of the video during editing. The incorporation of an additional appearance branch enables users to exert fine-grained control over the edited key frame. These two side branches seamlessly integrate into the main branch, which is constructed upon existing text-to-image (T2I) generation models, through learnable temporal layers. The versatility of our framework is demonstrated through a diverse range of choices in both structure representations and personalized T2I models, as well as the option to provide the edited key frame. To facilitate comprehensive evaluation, we introduce the \textit{BalanceCC} benchmark dataset, comprising 100 videos and 4 target prompts for each video. Our extensive user studies compare CCEdit with eight state-of-the-art video editing methods. The outcomes demonstrate CCEdit's substantial superiority over all other methods. 
\end{abstract}

\let\thefootnote\relax\footnotetext{$\affmark[1]$~CCEdit is currently a research project, and there are no immediate intentions to integrate it into a product or extend public accessibility. Any future research endeavor will adhere to Microsoft's AI principles.}

%% file: sec/1_intro.tex
\section{Introduction}
In recent years, the domain of visual content creation and editing has undergone a profound transformation, driven by the emergence of diffusion-based generative models \cite{ho2020denoising,song2020score,dhariwal2021diffusion}. A large body of prior research has demonstrated the exceptional capabilities of diffusion models in generating diverse and high-quality images \cite{rombach2022high,ramesh2022hierarchical,saharia2022photorealistic} and videos \cite{ho2022imagen,singer2022make,blattmann2023align}, conditioned by text prompts. These advancements have naturally paved the way for innovations in generative video editing \cite{xing2023simda,chai2023stablevideo, wang2023edit, ouyang2023codef,yang2023rerender,zhao2023controlvideo,qi2023fatezero,liu2023dynvideo}. 

Generative video editing, despite its rapid advancement, continues to face a series of significant challenges. 
These challenges include accommodating diverse editing requests, achieving fine-grained control over the editing process, and harnessing the creative potential of generative models. Diverse editing requirements include tasks such as stylistic alterations, foreground replacements, and background modifications. Generative models, while powerful and creative, may not always align perfectly with the editor's intentions or artistic vision, resulting in a lack of precise control. 
In response to these challenges, this paper introduces CCEdit, a versatile generative video editing framework meticulously designed to strike a harmonious balance between controllability and creativity while accommodating a wide range of editing requirements. 

CCEdit achieves its goal by effectively decoupling structure and appearance control in a unified \textit{trident network}. This network comprises three essential components: the main text-to-video generation branch and two accompanying side branches dedicated to structure and appearance manipulation. The \textit{main branch} leverages a pre-trained text-to-image (T2I) diffusion model \cite{rombach2022high}, which is transformed into a text-to-video (T2V) model through the insertion of temporal modules. The \textit{structure branch}, implemented as ControlNet \cite{zhang2023adding}, is responsible for digesting the structural information extracted from each frame of the input video and seamlessly infusing it into the main branch. Simultaneously, the \textit{appearance branch} introduces an innovative mechanism for precise appearance control, when an edited reference frame is available. The structure and appearance branches are effectively integrated into the central branch through learnable temporal layers. These layers serve not only as a cohesive link, aggregating information from side branches, but also as a crucial element ensuring temporal consistency across the generated video frames. 

In highlighting the versatility of our framework, we provide a wide range of control choices for both structure and appearance manipulation. For structure control, users can choose from various types of structural information, including line drawings~\cite{chan2022learning}, PiDi boundaries~\cite{su2021pixel}, and depth maps~\cite{ranftl2020towards}, all of which can serve as input to the structure branch. On the appearance control front, the main branch already provides an inherent mechanism, allowing control through text prompts. 
Additionally, personalized T2I models from the Stable Diffusion community, such as DreamBooth and LoRA~\cite{ruiz2023dreambooth,hu2021lora}, can be integrated as plugins into CCEdit, offering greater flexibility and creativity.
More importantly, the appearance branch can accommodate the referenced key frame, facilitating fine-grained appearance control. Notably, all these control options are seamlessly integrated within the same framework, yielding editing outcomes that demonstrate both temporal coherence and precision. This not only underscores the versatility of our solution but also ensures ease of adoption, making it a compelling choice for AI-assisted video editing. 

To address the challenges inherent in evaluating generative video editing methods, we introduce the \textit{BalanceCC benchmark} dataset. Comprising 100 diverse videos and 4 target prompts for each video, this dataset includes detailed scene descriptions and attributes related to video category, scene complexity, motion, among others. These descriptions are generated with the assistance of the cutting-edge GPT-4V(ision) model~\cite{openai2023gpt4,gpt4v,gpt4vcontribution,gpt4vblog} and then refined by human annotators. Through extensive experimental evaluations on this dataset, we not only confirm the outstanding functionality and editing capabilities of CCEdit, but also underscore the comprehensiveness of the benchmark dataset. We firmly believe that BalanceCC stands as a robust and all-encompassing evaluation platform for the dynamic field of generative video editing.

%% file: sec/2_related_work.tex
\section{Related Work}
\subsection{Diffusion-based Image and Video Generation}
Diffusion models (DM)~\cite{ho2020denoising,song2020score,dhariwal2021diffusion} have demonstrated exceptional capabilities in the field of image synthesis. 
These models indeed help by learning to approximate a data distribution through the iterative denoising of a diffused input.
What makes DMs truly practical is the incorporation of text prompt as condition to control the output image during the generative process~\cite{nichol2022glide,ramesh2021zero,saharia2022photorealistic,rombach2022high}. 
Apart from the proliferation of advanced techniques in the field of image synthesis, DMs have also excelled in video generation \cite{ho2022imagen,blattmann2023align,singer2022make,nichol2022glide}. 
This is achieved by integrating modulated spatial-temporal modules, enabling the synthesis of high-quality videos while maintaining temporal consistency.

\begin{figure*}[htbp]
\vspace{-5pt}
\centerline{\includegraphics[width=1.0\linewidth]{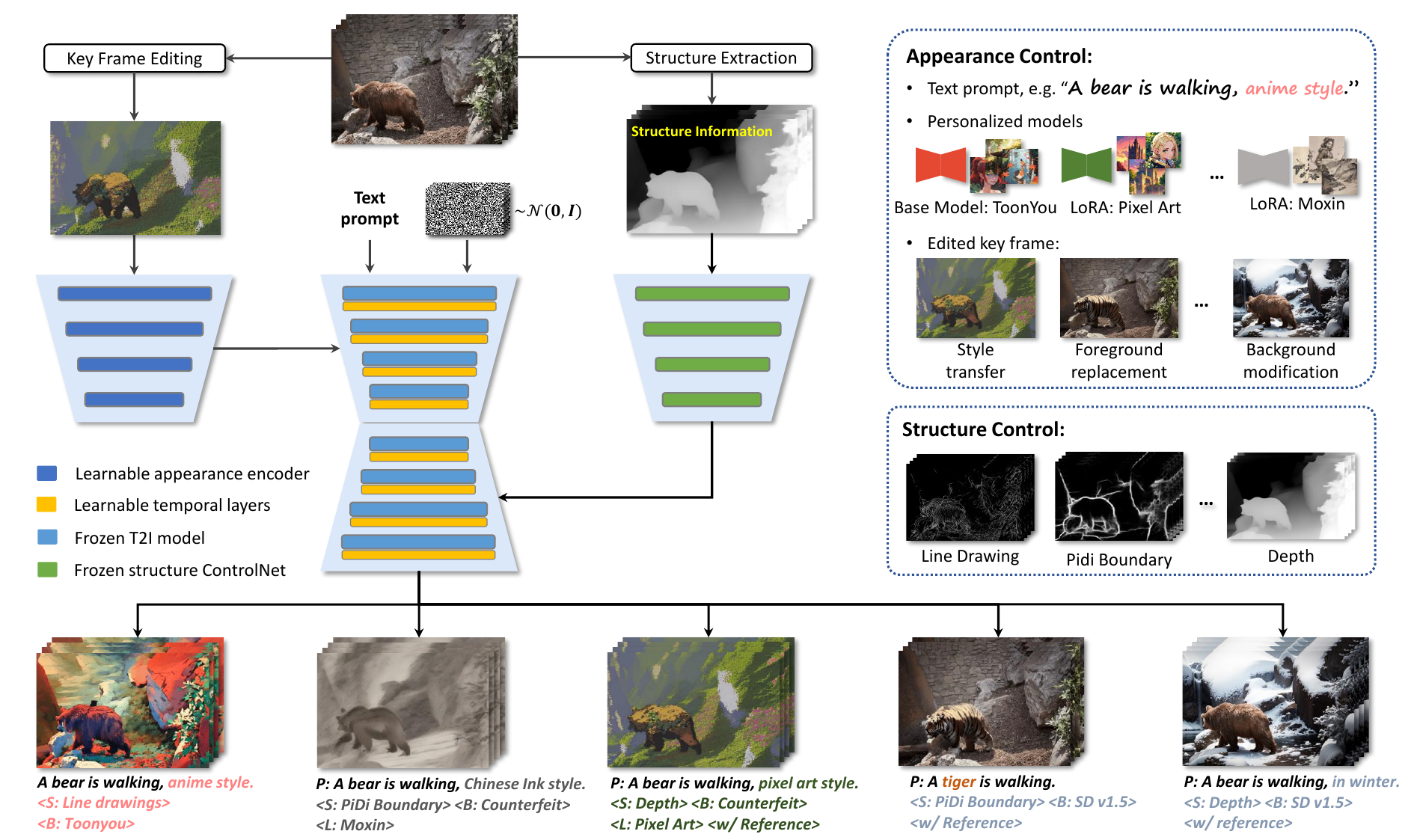}}
    \vspace{-3pt}
    \caption{
    \textbf{Illustration of our overall framework.} 
    Structure and appearance information in the target video are modulated independently, and seamlessly integrated into the main branch. 
    Structure control is conducted via the pre-trained ControlNet~\cite{zhang2023adding}.
    Appearance control is achieved precisely by the edited key frame. 
    Details regarding the autoencoder and iterative denoising process are omitted for simplicity. ``\textbf{P}'', ``\textbf{S}'', ``\textbf{B}'', ``\textbf{L}'' indicate prompt, structure, base model, and LoRA, respectively.
    }
    \vspace{-10pt}
\label{fig:overview}
\end{figure*}

\subsection{Video Editing with Diffusion Models}
Recent studies leverage the inherent generative priors of DMs for image editing \cite{meng2021sdedit,couairon2022diffedit,avrahami2022blended,tumanyan2023plug,parmar2023zero,hertz2022prompt}.
The same idea is also applied in the field of video editing. 
Unlike image editing, video editing involves not only the manipulation of appearance-based attributes but also requires the meticulous preservation of temporal coherence throughout frames.
A lapse in maintaining this temporal coherence can result in visual artifacts, such as flickering and degradation.

Some generative video editing methods \cite{zhang2023controlvideo,yang2023rerender,qi2023fatezero,wang2023zero,ceylan2023pix2video,khachatryan2023text2video,geyer2023tokenflow} strive to achieve training-free temporal consistency. They accomplish this by transitioning from spatial self-attention mechanisms within T2I diffusion models to temporal-aware cross-frame attention techniques.
Some other methods \cite{wu2023tune,shin2023edit,liu2023video,zhao2023controlvideo} perform per-video fine-tuning. They focus on optimizing the parameters of pre-trained T2I models according to the input video, aiming to achieve temporal coherence within the target video.
However, this optimization for each input video can be time-consuming and inadequate tuning of the temporal modules might lead to suboptimal temporal coherence. 
Recent studies \cite{xing2023simda,guo2023animatediff,liew2023magicedit} have introduced trainable temporal layers to construct T2V generative models. These models are trained on extensive text-video paired datasets, and they are used in both video generation and editing tasks \cite{esser2023structure,molad2023dreamix}.

Unlike previous work, this study does not seek a simple fix to existing T2I models for video editing, nor does it attempt to train a full-fledged T2V model. Instead, we introduce a unique network architecture tailored for video editing. Our approach involves dataset-level fine-tuning, circumvents the expenses associated with per-video tuning during inference time, and prioritizing the effective training of temporal layers to achieve robust model performance. 



%% file: sec/3_approach.tex
\section{Approach}
\subsection{Preliminary}
\myparagraph{Diffusion models} \cite{ho2020denoising} are probabilistic generative models that approximate a data distribution $p(\mathbf{x})$ by gradually denoising a normally distributed variable. 
Specifically, DMs aim to learn the reverse dynamics of a predetermined Markov chain with a fixed length of $T$.
The forward Markov chain can be conceptualized as a procedure of injecting noise into a pristine image.
Empirically, DMs can be interpreted as an equally weighted sequence of denoising autoencoders ${\epsilon}_{\theta}(\mathbf{x}_t,t)$ where $t={1,...,T}$.
These autoencoders are trained to predict a denoised variant of the noisy input $\mathbf{x}_t$. 
The corresponding objective can be simplified to
\begin{equation}
\vspace{-3pt}
\label{eq:diffusion_optimization_objective}
    \mathbb{E}_{\mathbf{x}_0,t,\epsilon \sim \mathcal{N}(\mathbf{0},\mathbf{I})} [{\| \epsilon-{\epsilon}_{\theta}(\mathbf{x}_{t},t) \|}_{2}^{2} ].
\vspace{-3pt}
\end{equation}

\myparagraph{Latent diffusion models} (LDMs) are trained in the learned latent representation space.
The bridge between this latent space and the original pixel-level domain is established via a perceptual compression model. 
The perceptual compression model is composed of an encoder $\mathcal{E}$ and a decoder $\mathcal{D}$, where $\mathbf{z}=\mathcal{E}(\mathbf{x})$ and $\mathbf{x} \approx \mathcal{D}(\mathcal{E}(\mathbf{x}))$. 
Then the optimization objective in Eq. (\ref{eq:diffusion_optimization_objective}) is modified as
\begin{equation}
\vspace{-3pt}
    \mathbb{E}_{\mathbf{z}_0,t,\epsilon \sim \mathcal{N}(\mathbf{0},\mathbf{I})} [{\| \epsilon-{\epsilon}_{\theta}(\mathbf{z}_{t},t) \|}_{2}^{2} ].
\vspace{-3pt}
\end{equation}


\subsection{The CCEdit Framework}
\label{sec:overall_framework}

The primary objective of our work is to empower creative control in video editing. Although creativity naturally emerges in generative models, achieving controllability is a more complex endeavor. To address this challenge, CCEdit strategically decouples the management of structure and appearance within a unified trident network. In Fig. \ref{fig:overview}, we provide an illustrative overview of the framework's architecture, which comprises three vital components. 


\myparagraph{The main branch.} The main branch of our model fundamentally operates as a text-to-video generation network. It is built upon the well-established text-to-image model, Stable Diffusion \cite{rombach2022high}. We transform this model into a text-to-video variant by incorporating temporal layers into spatial layers of both the encoder and decoder. 
This entails the addition of a one-dimensional \textit{temporal layer} with the same type as its previous \textit{spatial layer}, \ie, convolution blocks and attention blocks.
Besides, we also use the skip connection and zero-initialized \textit{projection out layer} of each newly added temporal layer for stable and progressive updating, which has been proven to be effective~\cite{zhang2023adding,singer2022make,guo2023animatediff}.
The zero-initialized projection out layer is instantiated as a linear layer.
Formally, let $\mathcal{F}(\cdot;{\Theta}_{s})$ be the 2D spatial block, $\mathcal{F}(\cdot;{\Theta}_{t})$ be the 1D temporal block, and $\mathcal{Z}(\cdot;{\Theta}_{z})$ be the zero-initialized projection out layer, where ${\Theta}_{s}$, ${\Theta}_{t}$, and ${\Theta}_{z}$ represent corresponding network parameters. The complete process of one pseudo-3D block that maps the input feature $\mathbf{u}$ to the output feature $\mathbf{v}$ is written as
\begin{equation}
\vspace{-2pt}
		\mathbf{v}=\mathcal{F}(\mathbf{u};{\Theta}_{s}) + \mathcal{Z}(\mathcal{F}(\mathcal{F}(\mathbf{u};{\Theta}_{s});{\Theta}_{t});{\Theta}_{z}),
\label{eq:main_branch}
\end{equation}
where $\mathbf{u}$ and $\mathbf{v}$ are both 3D feature maps, \ie, $\mathbf{u}\in\mathbb{R}^{l\times h \times w \times c}$ with $\{l, h, w, c\}$ as the number of frames, height, width, and the number of channels, respectively.

Moreover, we draw inspiration from AnimateDiff~\cite{guo2023animatediff} and VideoLDM~\cite{blattmann2023align}, which advocates the shared utilization of temporal layers among personalized T2I models such as DreamBooth \cite{ruiz2023dreambooth} and LoRA \cite{hu2021lora}.
The key aspect of it is training the temporal layers while keeping the spatial weights frozen.
We follow this schedule to inherit the T2I model's compatibility and visual generation capability.

\myparagraph{The structure branch.} The introduction of the structure branch is motivated by the common need in video editing tasks to preserve frame structure for non-edited or style-transferred segments. Striking a delicate balance between maintaining faithful frame structure and allowing the generative model ample creative freedom poses a significant challenge. The structure branch is implemented with the pre-trained ControlNet~\cite{zhang2023adding}. To accommodate varying levels of structure control, we use various types of structure representation, including line drawings~\cite{chan2022learning}, PiDi boundaries~\cite{su2021pixel}, and depth maps~\cite{ranftl2020towards}, ensuring adaptability to control structure at different degrees.

Specifically, the structure representation from all frames is extracted individually and injected into the main branch. Each frame undergoes preprocessing to derive a structure representation, and the weights of the ControlNet are held in a frozen state during training, emphasizing the preservation of learned structural features. 
Formally, let $\mathcal{F}(\cdot;{\Phi}_c)$ denote the ControlNet that maps structure information into features, and $\mathcal{Z}(\cdot;{\Phi}_{z1})$ and $\mathcal{Z}(\cdot;{\Phi}_{z2})$ denote the two instances of zero convolutions in \cite{zhang2023adding}.
Then the process of adding structure control to the 3D-aware feature $\mathbf{v}$ is
\begin{equation}
\vspace{-2pt}
    \mathbf{v}_s = \mathbf{v} + \mathcal{Z}(\mathcal{F}(\mathbf{z}_t + \mathcal{Z}(\mathbf{c}_s;{\Phi}_{z1}); {\Phi}_c); {\Phi}_{z2}),
\label{eq:structure_control}
\end{equation}
where $\mathbf{z}_t$ denotes the noisy input in latent space, $\mathbf{c}_s$ denotes the structure condition of the video sequence, and $\mathbf{v}_s$ denotes the feature aware of structure information.

\myparagraph{The appearance branch.} 
In addition to using text prompts and incorporating personalized models for appearance control, we introduce a novel design—the appearance branch. This architectural innovation introduces a pioneering approach for fine-grained appearance control, allowing for the integration of an edited frame as a detailed reference in the context of video editing. 
Since the editing of key frame can be accomplished through precise user edits or by using advanced off-the-shelf image editing algorithms, the introduction of appearance branch provides our framework with greater creativity and controllability.
Specifically, a key frame is initially assigned to the latent variable by the encoder $\mathcal{E}$.
Subsequently, a neural network with similar architecture to the main branch's encoder extracts multi-scale features.
The extracted features are incorporated into the main branch. 
Through this design, the appearance information from the edited key frame propagates to all frames via the temporal modules, effectively achieving the desired creative control in the output video.
Formally, suppose $\mathcal{F}(\cdot;\Psi)$ is the encoder that maps the pixel-wise appearance of the key frame into features, $\mathcal{Z}(\cdot;\Psi_{z})$ denotes the zero convolution projection out layer, $\mathbf{v}^{j}$ indicates the feature of the j-\textit{th} frame, and $\mathbf{c}_{a}^{j}$ is the key frame. Then the process of adding appearance control to the features is as follows
\begin{equation}
\vspace{-2pt}
    \mathbf{v}_{a}^{j} = \mathbf{v}^j + \mathcal{Z}(\mathcal{F}(\mathcal{E}(\mathbf{c}^{j}_{a});\Psi);\Psi_{z})
    \label{eq:appearance_branch},
\end{equation}
where $\mathbf{v}_{a}^{j}$ is the j-\textit{th} feature, aware of the edited appearance.



\myparagraph{Training.}
Before training, we initialize the spatial weights of the main branch with pre-trained T2I models. Temporal weights are randomly initialized while the projection out layers are zero-initialized.
We instantiate the model in the structure branch by pre-trained ControlNets~\cite{zhang2023adding}.
As for the appearance branch, we copy the encoder of pre-trained T2I model and remove text cross-attention layers.
During training, given the latent variables $\mathbf{z}_{0}=\mathcal{E}(\mathbf{x}_0)$ of an input video clip $\mathbf{x}_{0}$. Diffusion algorithms progressively add noise to it and produce the noisy input $\mathbf{z}_{t}$. 
Given conditions of time step $t$, text prompt $\mathbf{c}_{t}$, structure information $\mathbf{c}_{s}$, and appearance information $\mathbf{c}_{a}^j$ of the key frame, the overall optimization objective is 
\begin{equation}
\vspace{-2pt}
    \mathbb{E}_{\mathbf{z}_0,t,\mathbf{c}_{t},\mathbf{c}_{s},\mathbf{c}_{a}^j,\epsilon \sim \mathcal{N}(\mathbf{0},\mathbf{I})}[{\| \epsilon-{\epsilon}_{\theta}(\mathbf{z}_{t},t,\mathbf{c}_{t},\mathbf{c}_{s},\mathbf{c}_{a}^j) \|}_{2}^{2} ],
\end{equation}
where $\epsilon_{\theta}$ indicates the whole network to predict the noise added to the noisy input $\mathbf{z}_{t}$. 
We freeze the spatial weights in the main branch and the weights in the structure branch. Concurrently, we update the parameters of the newly incorporated temporal layers in the main branch, as well as the weights in the appearance branch. 
By default, the appearance branch takes the center frame of the video clip as input.

\myparagraph{Inference with anchor prior.} 
We find that, in some challenging cases, the edited video may exhibit large areas of flickering.
This is often caused by inconsistent structural representations extracted by image-level pre-processing modules.
Therefore, we propose a simple yet efficient strategy to improve the stability and quality of the result by modifying the start noise. 
Specifically, consider the individual noise sequence $[\epsilon_\text{ind}^1,...,\epsilon_\text{ind}^l]$ and the edited center frame $\mathbf{c}_a^j$, where $l$ and $j$ indicate the frame numbers and the index of the edited key frame, respectively.
The start noise ${{\epsilon}^i}$ for each frame is modified as
\begin{equation}
\vspace{-2pt}
\epsilon^i = {\epsilon}^{i}_\text{ind} + {\alpha}\mathcal{E}(\mathbf{c}_a^j), 
\label{eq:anchor_prior}
\end{equation}
where $\alpha$ is the hyperparameter that controls the strength of prior, and $\mathcal{E}(\mathbf{c}_a^j)$ is the latent of the edited key frame. 
We call this strategy \textit{anchor prior}, which is tailored for our pipeline of editing videos with an reference key frame.
We empirically found that $\alpha=0.03$ works well in most cases. 
The intuition behind it lies in that the video frames are usually similar to each other.
The operation of adding noise to diffusion models tends to rapidly destroy high-frequency information while slowly degrading low-frequency information. 
Therefore, the anchor prior can be seen as providing a bit of low-frequency information to all frames while ensuring that the distribution remains almost unchanged (achieved by small $\alpha$), thus becoming better starting points.

\subsection{Editing for Long Videos}
\label{sec:approach_longvideoediting}

Video editing tools face a challenge in maintaining a consistent look and feel across clips that span tens of seconds, equivalent to hundreds of frames. The inherent limitation of generative models, processing only a dozen frames per inference due to memory constraints, introduces variability in results, even with a fixed random seed. CCEdit addresses this challenge with its fine-grained appearance control, enabling the editing of long videos into a cohesive look and feel through extension and interpolation modes.

In essence, let $L+1$ represent the frames CCEdit processes in one run. For videos exceeding $L+1$ frames, we select one key frame for every $L$ frames. In the initial run, the first $L+1$ key frames undergo editing. Subsequent runs, in extension mode, treat the last edited frame from the previous run as the first frame. The edited result serves as a reference for the appearance branch. This process iterates until all key frames are processed. Transitioning to the interpolation mode, two adjacent frames become the first and last frames of an inference run to edit the $L-1$ intermediate frames, and both edited frames serve as references for the appearance branch. This continues until all frames are edited. This meticulous process ensures consistent editing results throughout the entire video.

%% file: sec/4_benchmark.tex
\section{BalanceCC Benchmark}
\subsection{Overview}
While generative video editing has gained considerable attention as a growing research field, the absence of a standardized benchmark for assessing the efficacy of different approaches poses a potential hindrance to the technical progression of the field. Despite the recent introduction of TGVE 2023~\cite{wu2023cvpr} as an evaluation benchmark, it is crucial to note that the videos within this benchmark present challenges such as severe camera shake, overly complex scenes, blur, and low frame rates. In response to this, we introduce \textit{BalanceCC}, a benchmark that contains 100 videos with varied attributes, designed to offer a comprehensive platform for evaluating video editing, focusing on both controllability and creativity.

\subsection{Benchmark Establishment}
We curated a collection of 100 open-license videos suitable for legal, non-stigmatizing modifications.
These videos range from 2 to 20 seconds in duration, each with a frame rate of about 30 fps.
Besides, we utilize GPT-4V(ision)~\cite{openai2023gpt4,gpt4v,gpt4vcontribution,gpt4vblog} as an assistant to establish this benchmark.
For each video, GPT-4V(ision) provides a description and assigns a complexity score to the scene using the center frame as a reference, with ratings from $1$ (Simple) to $3$ (Complex). Additionally, we manually annotate each video for camera movement, object movement, and categorical content, with motion rated on a scale from $1$ (Stationary) to $3$ (Quick), and categories that include humans, animals, objects, and landscapes. 
Following this, GPT-4V(ision) is tasked to craft target prompts for video editing, encompassing style, object, and background alterations, along with compound changes. This process, while akin to TGVE 2023 \cite{wu2023cvpr}, we additionally introduce a ``Fantasy Level'' to indicate the imaginative and creative degree of the target prompt. These measures are intended to assist researchers in appraising the applicability of various methods to source videos and in gauging their potential. 
See supplementary for details on the prompting pipeline, specific instructions, principles of labeling, and illustrative examples. 

\subsection{Statistics}
\begin{figure}[t]
    \vspace{-4pt}
\centerline{\includegraphics[width=1.0\linewidth]{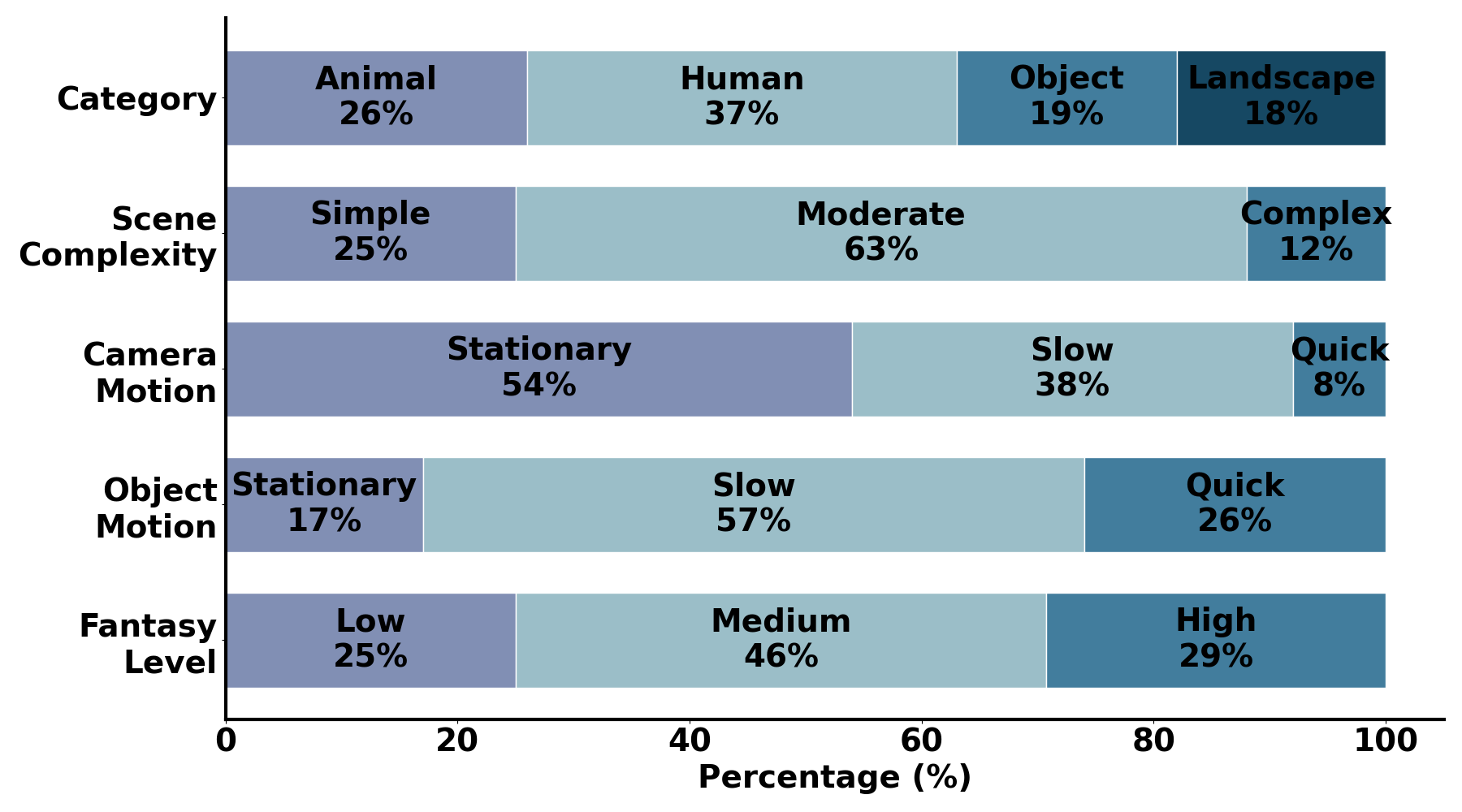}}
    \caption{
    Illustration of the statistics on BalanceCC.
    }
    \label{fig:benchmark_analysis}
    \vspace{-4pt}
\end{figure}

The overall distribution of BalanceCC is illustrated in Fig. \ref{fig:benchmark_analysis}. For the data of original videos, the distribution across categories tends towards uniformity, yet the ``Human'' category is slightly more prevalent than others. This was a deliberate choice, as editing human subjects is more practically significant and, due to the complexity of human and facial structures, editing in the ``Human'' category presents more challenges. Regarding ``Scene Complexity'' and ``Object Motion'', videos with moderate and slow levels are slightly more common. In terms of ``Camera Motion'', videos of lower levels predominate (Stationary: $54\%$, Slow: $38\%$). Finally, regarding the ``Fantasy Level'' distribution in target prompts, there is a relatively balanced allocation, with a marginal inclination towards videos categorized at a moderate level.

We hope that the aforementioned categorization of the benchmark will better assist researchers and users in understanding the strengths and weaknesses of a method, thus enabling targeted improvements and fostering rapid development in the field.

%% file: sec/5_experiments.tex
\section{Experiments}
\subsection{Implementation Details}
Stable Diffusion-v1.5 is used as the base T2I model in the main branch. We use the pre-trained ControlNet~\cite{zhang2023adding} for the structure information guidance. 
The training dataset combines WebVid-10M \cite{bain2021frozen} and a self-collected private dataset.
We trained the temporal consistency modules and appearance ControlNet towards various types of structural information, including line drawings~\cite{chan2022learning}, PiDi boundaries~\cite{su2021pixel}, depth maps detected by Midas~\cite{ranftl2020towards}, and human scribbles. 
Depth maps are used by default.
The control scales are set as $1$.
For the temporal interpolation model, we train it exclusively on depth maps, employing a smaller control scale of $0.5$. This approach is adopted because its requirement for structural information is comparatively less than that of other models.
During the training process, we first resize the shorter side to $384$ pixels, followed by a random crop to obtain video clips with a size of $384\times576$.
$17$ frames at $4$ fps are sampled from each video.
The batch size is $32$ and the learning rate is $3e-5$. We train each model for $100$K iterations.
During inference, we employ the DDIM~\cite{song2020denoising} sampler with $30$ steps, classifier-free guidance~\cite{ho2021classifier} of magnitude $9$.

\subsection{Applications}

\begin{figure}[t]
\small
\centerline{\includegraphics[width=1.0\linewidth]{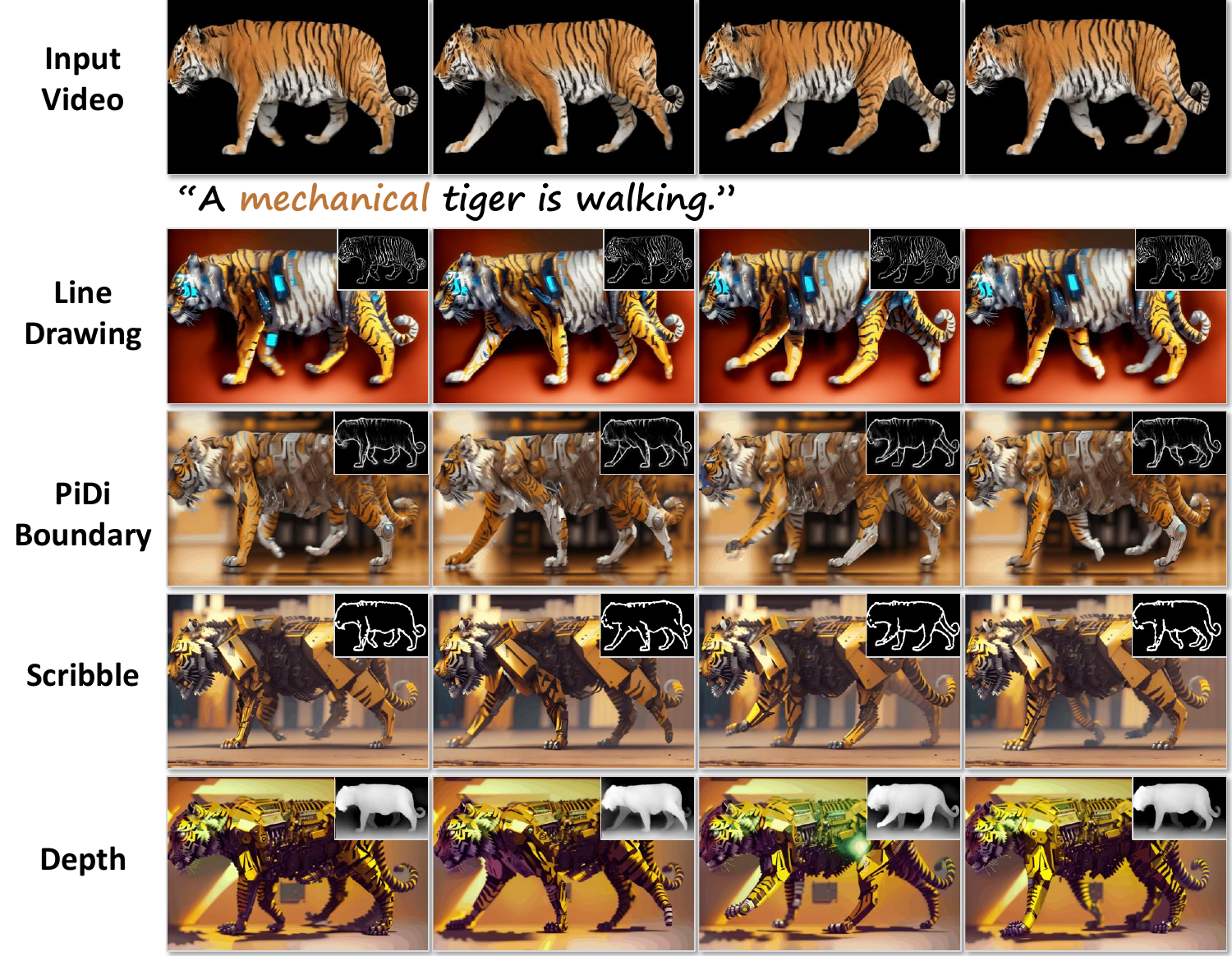}}
    \caption{
    \textbf{Results under different structural guidance.} 
    }
    \label{fig:exp_different_structural_control}
\end{figure}

\begin{figure}[t]
\centerline{\includegraphics[width=1.0\linewidth]{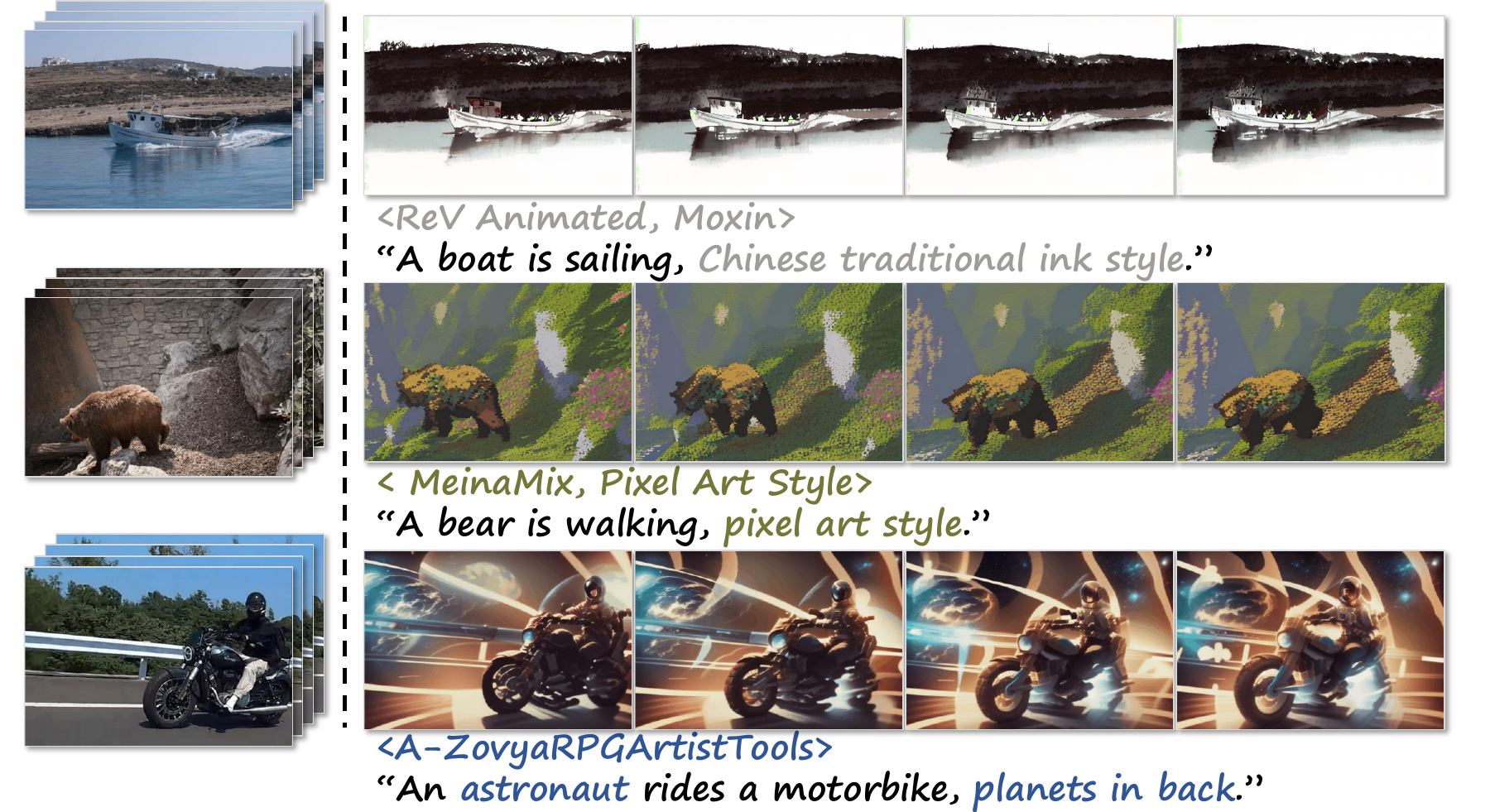}}
    \caption{\textbf{Results of video style translation.} 
    $\langle \cdot \rangle$ indicate the personalized T2I model we used.
    }
    \label{fig:exp_style_translation}
\end{figure}

\myparagraph{Controllable and creative style transfer.}
In CCEdit, the controllability and creativity of video style transfer are manifested in various dimensions. Two basic aspects include the diversity of structural information and the availability of off-the-shelf personalized models~\cite{huggingface,civitai}. 
The former enables users to customize the granularity and type of structural information retained from the original video, as depicted in Fig. \ref{fig:exp_different_structural_control}. 
The latter allows users to edit the video into their desired domain, as shown in Fig. \ref{fig:exp_style_translation}.

\begin{figure}[t]
\centerline{\includegraphics[width=1.0\linewidth]{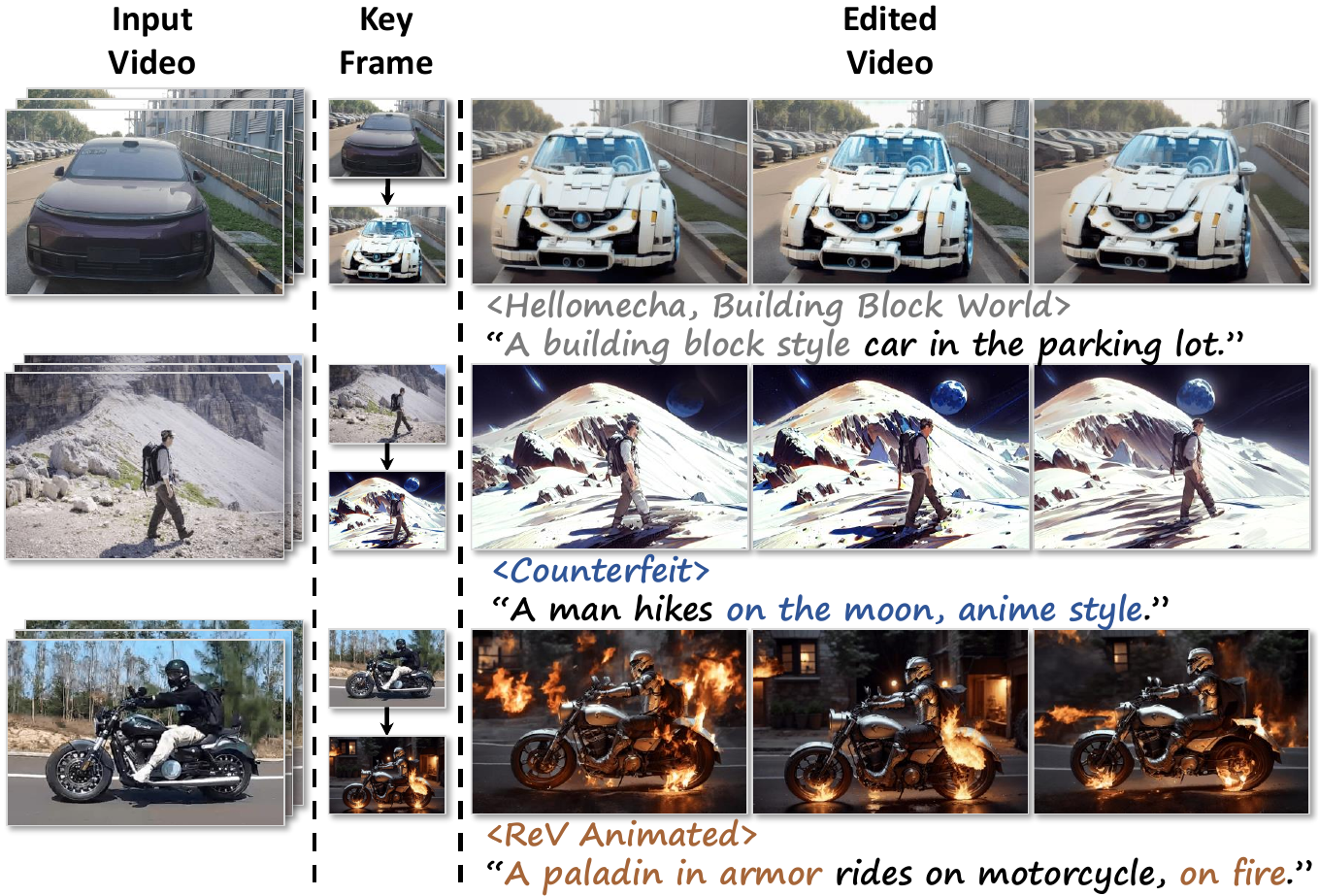}}
    \caption{\textbf{Video editing results with customized center frame as reference.} 
    The first row corresponds to customizing foreground, the second row corresponds to customizing background, and the third row is taking given reference image to affect the entire picture.
    $\langle \cdot \rangle$ indicate the personalized T2I model we used.
    }
\label{fig:exp_customizededit}
\end{figure}

\myparagraph{Video editing with precise appearance control.}
Sometimes, users require stronger control over the content they want to generate. For example, they may want to change only the foreground, alter just the background, or edit the texture content of a video in a specific way. 
Therefore, CCEdit focuses more on precise appearance control by initially modifying the key frame with image editing techniques and then using it as a reference for the entire video.
As depicted in Fig. \ref{fig:exp_customizededit}, we first edit the center frames of the videos by Stable Diffusion Web UI~\cite{AUTOMATIC1111_Stable_Diffusion_Web_2022}, followed by utilizing these edited center frames as guides for the video editing process. 
Thanks to end-to-end network training, our method coherently propagates edits from the key frame throughout the entire video.

\begin{figure}[t]
\vspace{-5pt}
\centerline{\includegraphics[width=1.0\linewidth]{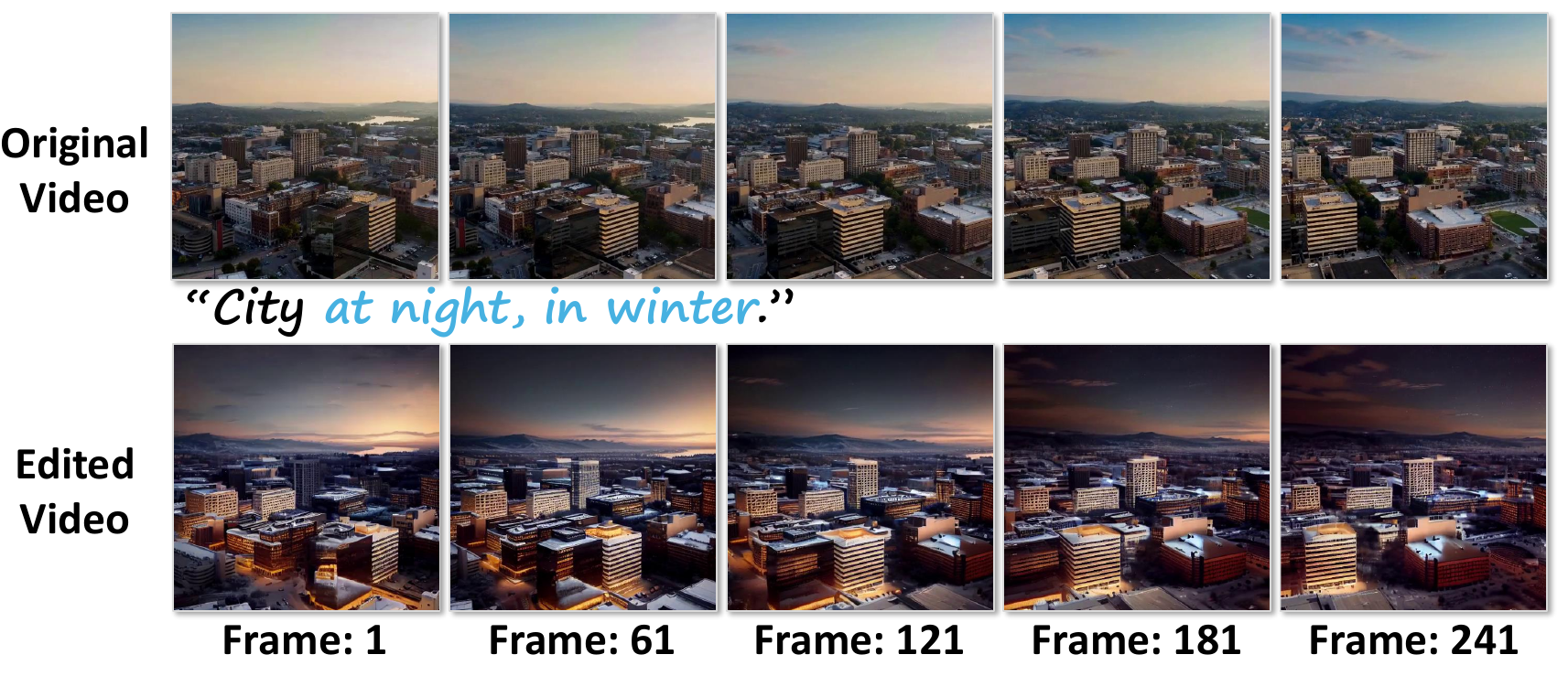}}
\vspace{-3pt}
    \caption{
    \textbf{Illustration of long video editing.} CCEdit achieves good consistency across over 240 frames. Zoom in for best view.
    }
    \label{fig:FigureExpVeryLongVideo}
\vspace{-5pt}
\end{figure}

\myparagraph{Long video editing.}
A seamless and visually appealing video typically necessitates a higher frame count and increased frame rate, elements that have been inadequately addressed by many contemporary video editing methodologies. CCEdit effectively resolves this through its hierarchical design for key frames editing, combined with iterative extension and a tailored temporal interpolation mechanism. This approach enables the editing of videos comprising up to hundreds of frames with $24$ fps (frames per second). An example is shown in Fig. \ref{fig:FigureExpVeryLongVideo}.

\begin{figure}[t]
\centerline{\includegraphics[width=1.0\linewidth]{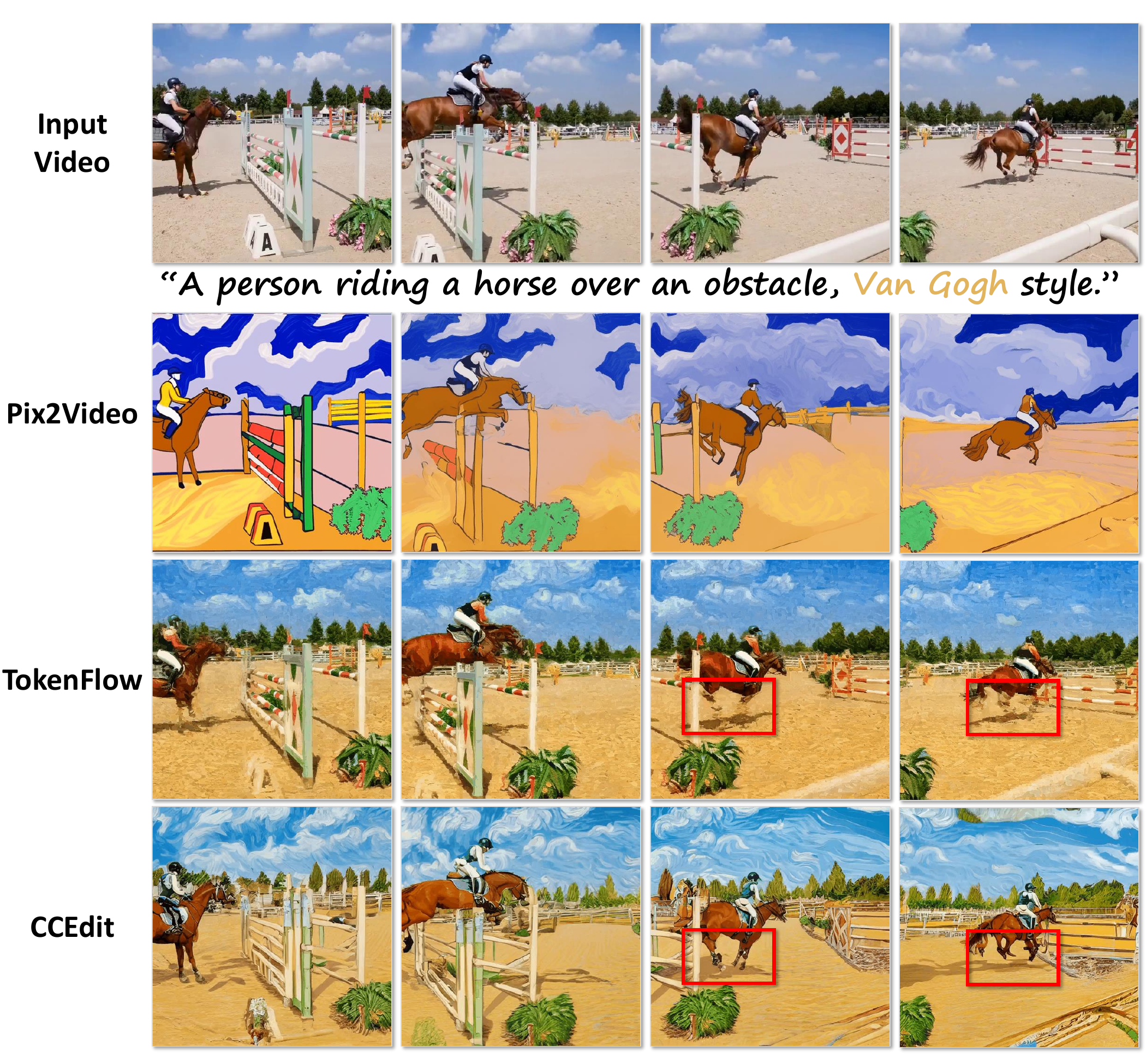}}
\vspace{-3pt}
    \caption{
    \textbf{Qualitative comparison results.} Red boxes reveals TokenFlow's inadequate local detail preservation, in contrast to our method's detailed, coherent output.
    Zoom in for best view.
    }
    \label{fig:exp_comparison}
    \vspace{-3pt}
\end{figure}

\subsection{State-of-the-Art Comparisons}
\label{sec:exp_sota_comparison}
\myparagraph{Datasets.}
We employ a smaller segment of our proposed benchmark, designated as \textit{mini-BalanceCC}. This subset encompasses $50$ videos, each randomly selected from the original BalanceCC dataset, ensuring a representative distribution similar to that of the original collection.

\myparagraph{Compared methods.}
To conduct an exhaustive comparison, we have selected eight representative video editing methodologies: Tune-A-Video~\cite{wu2023tune}, vid2vid-zero~\cite{wang2023zero}, Text2Video-zero~\cite{khachatryan2023text2video}, FateZero~\cite{qi2023fatezero}, Pix2Video~\cite{ceylan2023pix2video}, ControlVideo~\cite{zhang2023controlvideo}, Rerender A Video~\cite{yang2023rerender}, and TokenFlow~\cite{geyer2023tokenflow}.
Method details are omitted for brevity, and can be found in supplementary.
Regarding our approach, we employ depth maps as structure control.
For the appearance control, we adopt the off-the-shelf method of PnP-Diffusion~\cite{tumanyan2023plug} with the same hyper-parameters to automatically edit the center frame of each video clip. 
To ensure fairness in comparison, Stable Diffusion-v1.5 is used as the base model for all methods.

\myparagraph{Evaluation metrics.} 
In our preliminary study, we observed that automatic metrics, such as CLIP-Score \cite{hessel2021clipscore} to assess text alignment and frame consistency, do not fully align with human preferences~\cite{zhang2023towards,molad2023dreamix,wu2023cvpr}. 
We focused on collecting human preferences for a comprehensive user study, comparing our method against recent state-of-the-art techniques based on mean opinion score (MOS) and direct comparisons.
We gathered 1,119 scoring results from 33 volunteers, each reflecting all indicators for an edited video.
For automatic metric results, refer to the supplementary.


\myparagraph{Results.}
As illustrated in Tab. \ref{table:Quantitative_baseline}, CCEdit excels in both editing accuracy and aesthetic quality, and is just slightly inferior to TokenFlow in temporal smoothness. For overall impression, our approach achieved a MOS of 3.87 on a scale from 1 to 5. Among the eight reference methods, TokenFlow performed closest to ours, with an overall MOS of 3.58. The remaining seven methods scored between 1.5 to 3.0 on the MOS scale. 
As for direct comparisons, our method outperforms all eight reference schemes significantly. While TokenFlow remains the closest competitor, our CCEdit prevails in 52.9\% of test cases against it, trails in 32.4\%, and ties in 14.7\% of cases.

Furthermore, Fig. \ref{fig:exp_comparison} presents the qualitative results of the top three finalists (CCEdit, TokenFlow~\cite{geyer2023tokenflow}, and Pix2Video~\cite{ceylan2023pix2video}). It shows that Pix2Video struggles to keep temporal coherence, while TokenFlow demonstrates noticeable blurring. 
In contrast, our method can accurately achieve the editing objective while maintaining the temporal coherence as well as the structure of the input video. 


\begin{table}[t]
\vspace{-3pt}
\footnotesize
  \centering
  \begin{tabular}{@{\hspace{5pt}}l|@{\hspace{5pt}}c@{\hspace{5pt}}c@{\hspace{5pt}}c@{\hspace{5pt}}c|@{\hspace{5pt}}c@{\hspace{5pt}}c@{\hspace{5pt}}c@{\hspace{5pt}}}
    \toprule
    Method & Edit & Aes. & Tem. & Ove. & Win & Tie & Lose \\
    \midrule
    
    Tune-A-Video~\cite{wu2023tune}        & 3.24 & 3.01 & 2.72 & 2.77 & 16.4 & 6.9 & 76.7 \\
    vid2vid-zero~\cite{wang2023zero}        & 3.00 & 2.38 & 2.11 & 2.35 & 10.6 & 4.6 & 84.8 \\
    Text2Video-Zero~\cite{khachatryan2023text2video}     & 2.07 & 1.43 & 1.41 & 1.48 & 16.5 & 1.3 & 86.2 \\
    FateZero~\cite{qi2023fatezero}            & 2.47 & 3.16 & 3.30 & 2.79 & 16.6 & 3.6 & 79.8 \\
    Pix2Video~\cite{ceylan2023pix2video}           & 3.68 & 2.97 & 2.80 & 2.97 & 29.9 & 5.2 & 64.9 \\
    ControlVideo~\cite{zhang2023controlvideo}        & 3.01 & 2.71 & 2.60 & 2.66 & 13.8 & 5.6 & 80.6 \\
    Rerender A Video~\cite{yang2023rerender}    & 2.40 & 2.69 & 2.82 & 2.50 & 11.1 & 0.0 & 88.9 \\
    TokenFlow~\cite{geyer2023tokenflow}           & 3.78 & 3.61 & \textbf{3.79} & 3.58 & 32.4 & 14.7 & 52.9 \\
    \midrule
    CCEdit (Ours)       & \textbf{4.06} & \textbf{4.00} & 3.74 & \textbf{3.87} & - & - & - \\
    \bottomrule
  \end{tabular}
  \vspace{-2pt}
  \caption{
  \textbf{Left: Mean opinion scores (MOS) over different aspects of the generated video,}  including editing accuracy (Edit), aesthetics (Aes.), temporal consistency (Tem.), and overall impression (Ove.). Scores range from 1 to 5. \textbf{Right: Win, Tie, and Lose percentage in side-by-side comparisons with CCEdit. }
  }
  \label{table:Quantitative_baseline}
    \vspace{-3pt}
\end{table}

\vspace{-1pt}
\subsection{Ablation Study}
\vspace{-1pt}

\begin{figure}[t]
\vspace{-3pt}
\centerline{\includegraphics[width=1.0\linewidth]{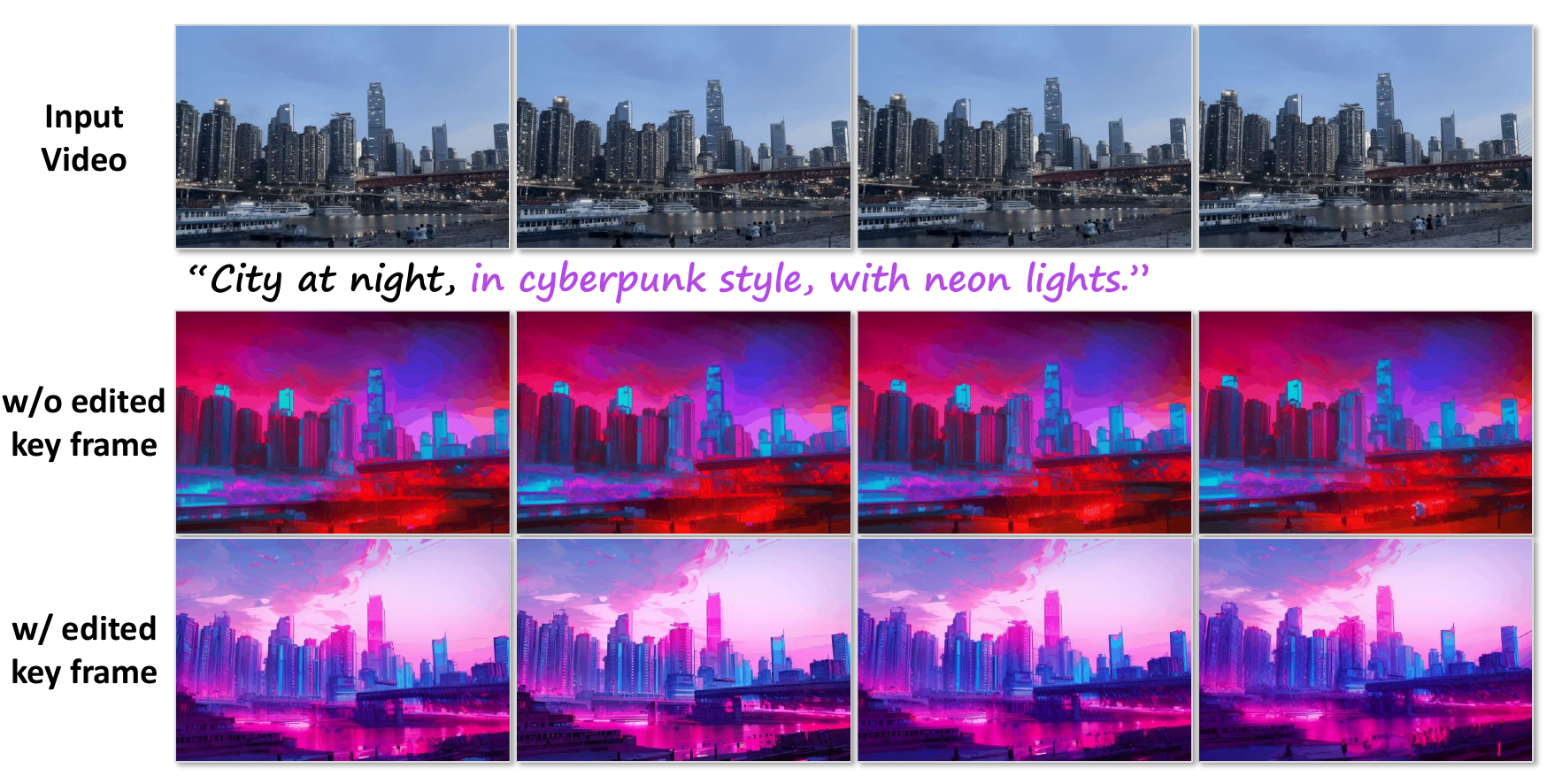}}
  \vspace{-3pt}
    \caption{
    \textbf{Ablation study on appearance control.} In some challenging cases, appearance control is crucial to achieving the expected results.}
    \label{fig:exp_different_appearance_control}
    \vspace{-3pt}
\end{figure}

\myparagraph{Appearance control.}
Fig. \ref{fig:exp_different_appearance_control} illustrates the importance of taking the edited key frame as a reference in certain scenarios. 
Initially, translating video scenes into ``\texttt{cyberpunk}'' style (1st row) solely through prompt adjustments appears challenging, as this word is unfamiliar to the pre-trained T2I model weights and the temporal consistency modules. 
Providing a customized center frame allows the network to smoothly extend its appearance to adjacent frames, creating a cohesive video.
Besides, we replicated the user study pipeline from Sec. \ref{sec:exp_sota_comparison} to evaluate the effectiveness of appearance control. The model without appearance control received a mean opinion score (MOS) of 2.88, significantly lower than the 3.87 scored by the process of editing one key frame first and then propagating to surrounding frames.

\begin{figure}[t]
\centerline{\includegraphics[width=1.0\linewidth]{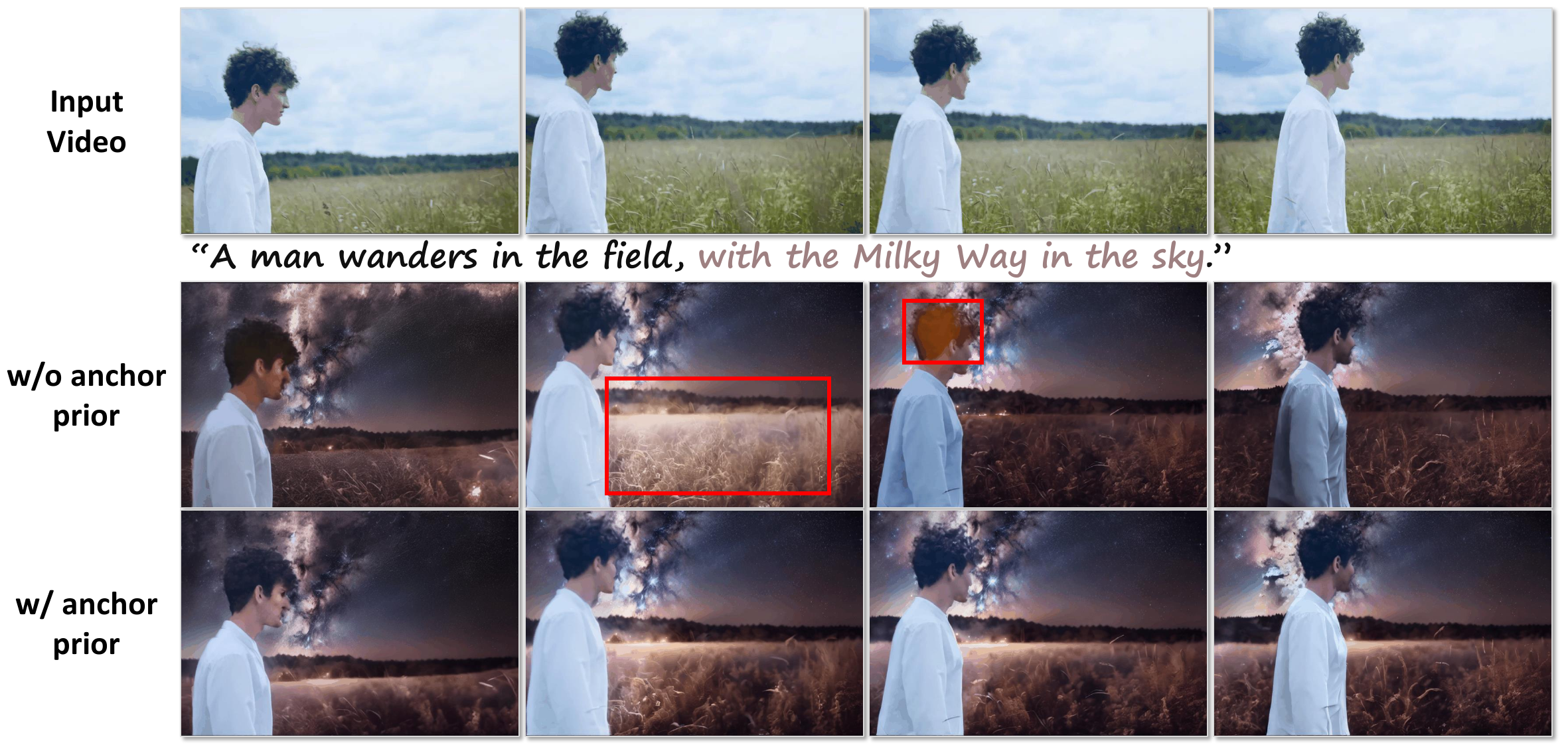}}
  \vspace{-3pt}
    \caption{
    \textbf{Ablation study on anchor prior.} Our proposed anchor prior helps a lot in stabilizing the appearance across frames. The red boxes demonstrate the localized flickering in the frames.
    }
    \label{fig:AblationAnchorPrior}
    \vspace{-3pt}
\end{figure}

\myparagraph{Anchor prior.}
Fig. \ref{fig:AblationAnchorPrior} demonstrates the ablation study for our anchor prior. It reveals that the absence of the anchor prior may lead to regional flickering in the video sequence, while its presence effectively mitigates this issue.

%% file: sec/6_limitation.tex
\section{Limitation and Future Works}
In our approach, structural control is exerted by explicitly extracting the structural representation from the source video and sustaining it via the structure branch. However, it may encounter challenges when tasked with substantial structural alterations-exemplified by the conversion of a ``cute rabbit'' into a ``majestic tiger.'' Addressing these complexities will be a primary objective of our future work.

%% file: sec/7_conclusion.tex
\section{Conclusion}
This paper presents an innovative trident network architecture specifically designed for generative video editing. 
This unified framework enables precise and controllable video editing while broadening creative possibilities. 
To address the challenges in evaluating generative video editing approaches, we introduce the meticulously curated BalanceCC benchmark dataset. Our aim is to pave the way for researchers in the generative video editing domain and equip practitioners with indispensable tools for their creative workflows.


%% file: sec/X_suppl.tex
\clearpage
\maketitle
\thispagestyle{empty}
\appendix


\section{Details of the Trident Network}
\begin{figure}[t]
\centerline{\includegraphics[width=1.0\linewidth]{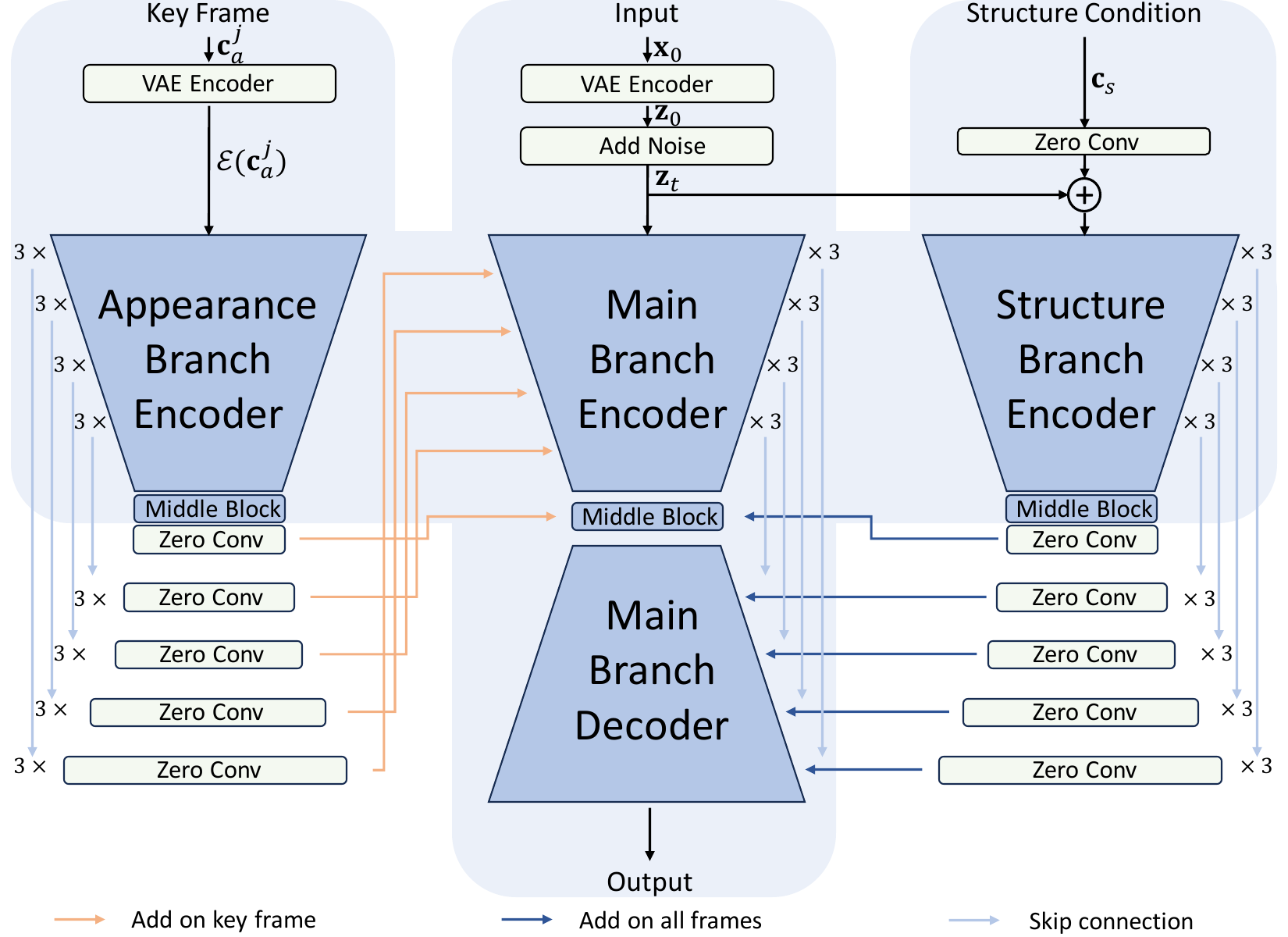}}
    \vspace{-2mm}
    \caption{\textbf{Illustration of our proposed trident network. Left: Appearance branch. Middle: Main branch. Right: Structure branch.} 
    Text prompts and time embedding are incorporated are omitted for simplicity.
    }\label{fig:reference_aware_control}
\end{figure}

The detailed architecture of our proposed \textit{trident network} is illustrated in Fig. \ref{fig:reference_aware_control}.
Specifically, in the appearance branch, the edited key frame $\mathbf{c}_a^j$ is encoded by the VAE encoder $\mathcal{E}$.
Then it's fed into the encoder of \textit{appearance branch}. 
Subsequently, the features extracted from each layer are fed into zero convolutions and the output are added to the corresponding features in the encoder side of \textit{main branch}. 
On the right side, \ie, the \textit{structure branch}, structure information $\mathbf{c}_s$ of original video clip is encoded by the zero convolution and fed into structure branch encoder. Similar to the appearance branch, features extracted are fed into zero convolutions.
Differently, the output are added to the corresponding features in the decoder side of main branch. 
The structure branch is instantiated by ControlNet~\cite{zhang2023adding}.
Note that in the original paper of ControlNet, it consists a tiny network to encode the pixel-wise structure representation. Here we omit it for simplicity.
Ultimately, the appearance information within the key frame is propagated to all frames through the temporal modules and the inherited structure information will ensure the structural fidelity, achieving the stable and controllable editing. 

It is important to highlight that, we don't use a train-from-scratch tiny encoder to encode the condition as ControlNet~\cite{zhang2023adding} does in the appearance branch.
Instead, we use the VAE encoder $\mathcal{E}$ to map the pixel-wise appearance into latent variable, which is in the same representation space as latent variable $\mathbf{z}_0$. 
The intuition behind is its inherent capacity to act as a natural bridge, mapping pixel-wise appearance into the latent space which is exactly the U-Net works in. 
Consequently, we are able to seamlessly copy the weights from the main branch encoder to initialize appearance branch, thereby accelerating and stabilizing the convergence process.

\section{BalanceCC Benchmark}
Our objective is to develop a benchmark dataset specifically designed for tasks involving controllable and creative video editing.
Therefore, we collected 100 open-license videos of different categories, including Animal, Human, Object, and Landscape. In addition, for each source video, we provided a text description and graded Camera Motion, Object Motion, and Scene Complexity on a scale from 1 to 3. For each video, there are four types of edits along with corresponding target prompts and Fantasy Levels (also ranging from 1 to 3), namely Style Change, Object Change, Background Change, and Compound Change. Our aim in doing so is to better compare the strengths and weaknesses of different methods and their areas of expertise, as well as to assist researchers in advancing their techniques.
In this section, we provide details about how to prompt GPT-4V(ision)~\cite{openai2023gpt4,gpt4v,gpt4vcontribution,gpt4vblog} to assistant us to establish our proposed BalanceCC benchmark and some illustrative examples. 
The BlanceCC benchmark will be public soon.
\subsection{Prompting Pipeline and Instructions}
GPT-4V(ision)~\cite{openai2023gpt4,gpt4v,gpt4vcontribution,gpt4vblog} is a multi-modal model that possesses powerful capabilities in visual understanding, language comprehension, conversational skills, and a vast repository of knowledge. Consequently, we aim to leverage these dual capabilities to help us establish the BalanceCC benchmark. The process is akin to seeking advice from a wise person with extensive knowledge and excellent vision. Specifically, we first inform GPT-4V(ision) of our intention to create a benchmark dataset dedicated to video editing, explaining our requirements including scene complexity, original prompts, target prompts, editing types, and the corresponding fantasy levels. Then, we send the center frame of each video clip to GPT-4V(ision), allowing it to output the content we need in the specified format. 
In our initial attempts, we observed that GPT-4V(ision) still experienced some hallucinations, overly detailed descriptions and expansions, and instances of forgetfulness during interactions. Consequently, we made repeated and emphasized adjustments in our prompt. Additionally, we found that merely describing our needs was insufficient to achieve the desired results. Our solution was to provide corresponding examples as references, which significantly improved the quality of the content provided by GPT-4V(ision).
The final prompt we used is as follows,
\begin{quotation}
Now I'm trying to build a benchmark for video editing. I need you to assist me in doing that. I will provide the center frame of each video for you. About the image, I hope you provide the following information to me:

  1. Classify this video into one of ``Human, Animal, Object, Landscape''.
  
  2. Describe this image, be brief, concise, and precise. Don't use too many adjectives.
  
  3. Try to generate four text prompts of different types to edit this video. Be creative and imaginative. Offer me the corresponding ``Editing Type'', ``Target Prompt'', and ``Fantasy Level'' of each prompt. The ``Editing Type'' should be one of ``Style Change, Object Change, Background Change, and Compound Change.'' About ``Style Change'', some examples are ``old movies'', ``impressionist style'', ``Van Gogh style'', ``neon lights style'', ``cyberpunk style'', ``sepia-toned photo'', ``grayscale'',  ``claymation style'', ``origami style'', ``oil painting style''. About ``Object Change'', just change the object into other ones, like ``dog to cat'', ``cat to tiger'', ``human to bear'', ``human to teddy bear'', and even some specific identities like ``Ironman''. About ``Background Change'', just change the background, here are some examples, ``in the Mars'', ``in the moon'', ``in the forest'', ``in the ocean'', ``in the castle''. You can pick one of the examples I provided, and I hope you can also consider other ones that you think are interesting or suit this video. About ``Compound Change'', just combine what mentioned above. Please remember, be creative and imaginative, and don't be too outrageous. Besides, all targets including ``Style Change, Object Change, Background Change, and Compound Change'' should be provided for one video. The form of ``Target Prompt'' should be just like a description of an video, don't say something like ``Transform the background into moon.'' Here is an example, the original prompt is ``A majestic black swan gracefully floats on calm waters, with its reflection visible.'', the ``Target Prompt'' can be ``An elegant flamingo swan gracefully floats on calm waters, with its reflection visible, set against a backdrop of a mystical enchanted forest.''. As for the ``Fantasy Level'' for each ``Target Prompt'', it indicates the degree of imagination. For example, if you change the cat to a tiger or change the background from autumn to winter, it can be seen as a relatively low degree of imagination. Transforming a cat into pixel tiger or tiger made of origami is relative high degree of imagination. Here is also 1-3 in total 3 levels. And similar to the description, be brief, concise, and precise. 
  
  4. Is the scene complex or not? Rank it from 1 to 3, corresponding to simple, moderate, and complex.
\end{quotation}

\subsection{Human Refinement}

Upon receiving initial outcomes from GPT-4V(ision), we engaged in a manual refinement and augmentation process. This primarily entailed the verification and rectification of existing annotations, along with the inclusion of additional details regarding the magnitude of camera and object motion within the video sequences. Specifically, our rule to define levels of different attributes is as follows:

\textit{Camera Motion}: 1 corresponds to stationary, indicating minimal scene change and camera movement. 2 corresponds to slow movement, where the camera moves steadily and slowly. 3 corresponds to scenarios with intense camera shake and rapid movement.

\textit{Object Motion}: 1 corresponds to stationary, where the target is almost motionless or has very minimal movement. 2 corresponds to slow movement, where the target follows a slow, simple, and regular trajectory (such as uniform linear motion). 3 corresponds to targets engaging in fast and complex movements (such as dancing and boxing).

\textit{Scene Complexity}: 1 corresponds to scenes with a single target and a clean background. 2 corresponds to scenes with a few targets where both the targets and the background are not complex. 3 corresponds to scenes with multiple foreground targets, complex backgrounds, and intricate depth relationships.

\textit{Fantasy Level}: 1 corresponds to simple target or background replacements and style transfers, such as transforming a dog into a cat or shifting to a Van Gogh painting style. 2 corresponds to more creative target and background replacements and style transfers, like replacing the background with a Martian landscape or turning an airplane into a dragon. 3 corresponds to complex and creative editing objectives combined together, with the Fantasy Level for Compound Change generally being 3.

\subsection{Illustrative Examples}
\begin{figure*}[t]
\centerline{\includegraphics[width=0.9\linewidth]{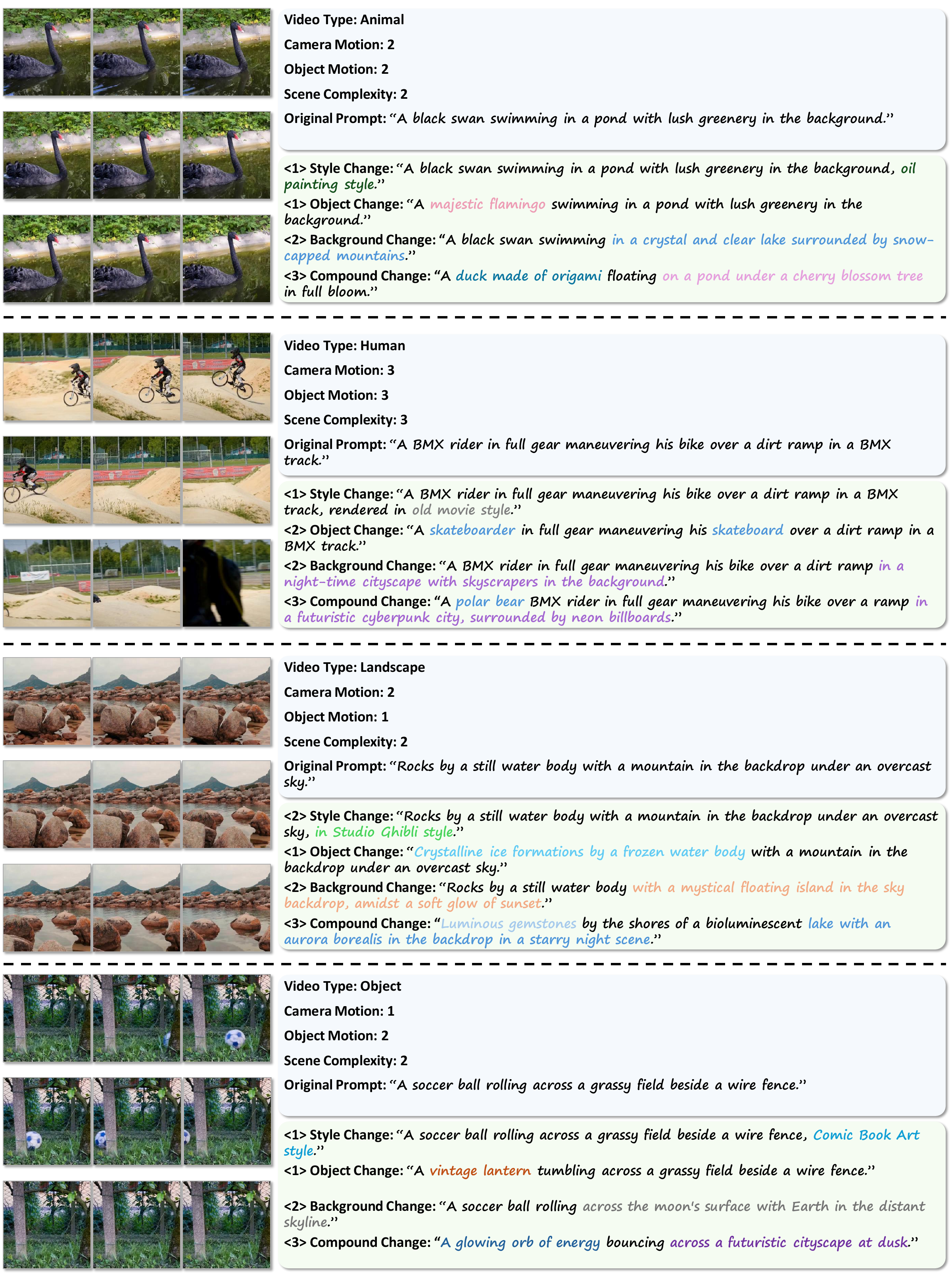}}
    \vspace{-6pt}
    \caption{\textbf{Illustrative examples of BalanceCC benchmark dataset.}
    }\label{fig:FigureSuppBenchmarkExamples}
\end{figure*}

Four illustrative examples are shown in Fig.~\ref{fig:FigureSuppBenchmarkExamples}.

\section{Experiments}



\subsection{Personalized T2I Models}
As mentioned in the main text, our method can integrate off-the-shelf personalized models as plugins, enabling the generation of domain-specific results. In this section, we briefly introduce the principles and specific implementations of personalized models. 

Stable Diffusion~\cite{rombach2022high} is trained on a huge dataset that encompasses a broad spectrum of domains~\cite{schuhmann2022laion}. Although the Stable Diffusion model is highly versatile and capable of generating a wide array of images, it occasionally falls short in specific details, particularly when it comes to generating human faces and hands, where subtle variations can markedly influence the overall perception.
Additionally, it often struggles to precisely meet users' expectations for specific content, styles, and attributes. 
Therefore, personalized T2I models are designed to address these challenges.
Two respective methods are DreamBooth~\cite{ruiz2023dreambooth} and LoRA~\cite{hu2021lora}. 
The former uses a unique string as an indicator to represent the corresponding domain or concept during training. Once trained, this indicator can be employed to transfer the expectations to the fine-tuned T2I model.
DreamBooth faces challenges due to the extensive weight parameters, making communication less convenient.
To use much less parameters and inherent the generalization of the base model, LoRA fine-tunes the model by preserving all original parameters and introducing the weight residuals $\Delta W$ to update the weights $W$. This process is formulated as $W' = W + \alpha \Delta W$, where $\alpha$ is the hyperparameter that controls the significance of the added $\Delta W$. 
Typically, the parameters of $\Delta W$ are significantly fewer than those of $W$.
Finally, two additional methods for creating robust personalized T2I base models are fine-tuning the entire model directly on the self-collected datasets and blending parameters from various models. 
Personalized T2I models play a crucial role in today's AI content generation. They empower both beginners and seasoned artists, as well as enthusiasts, to swiftly and autonomously produce stunning images and create new models.
A significant objectives of our framework is to ensure compatibility with personalized T2I models, allowing creators to freely combine and perform highly creative edits on videos using models from the community.

In this paper, we collect several personalized T2I base models and LoRA weights from CivitAI~\cite{civitai} and explored different combinations, which are illustrated in Table \ref{tab:personalized_models}.
Similar to previous work~\cite{guo2023animatediff}, we employ the ``trigger words'' to activate these personalized models. 
$\alpha$ of all LoRA models is set as $0.9$.

\begin{table}[t]
\small
  \centering
  \begin{tabular}{ccc}
    \toprule
    Model Name & Type \\
    \midrule
    Counterfeit & T2I Base Model \\
    ToonYou & T2I Base Model \\
    rev Animated & T2I Base Model \\
    HelloMecha & T2I Base Model \\
    hellonijicute25d & T2I Base Model \\
    A-Zovya Photoreal & LoRA \\
    kMechAnimal & LoRA \\
    Pixel Art Style & LoRA \\
    fat animal & LoRA \\
    Building Block World & LoRA \\
    MoXin & LoRA \\
    mechanical dog & LoRA \\
    \bottomrule
  \end{tabular}
  \caption{
  Personalized models utilized in this paper, all sourced from CivitAI~\cite{civitai}.
  }
  \label{tab:personalized_models}
\end{table}

\subsection{More Visualizations}
\begin{figure*}[t]
\centerline{\includegraphics[width=0.85\linewidth]{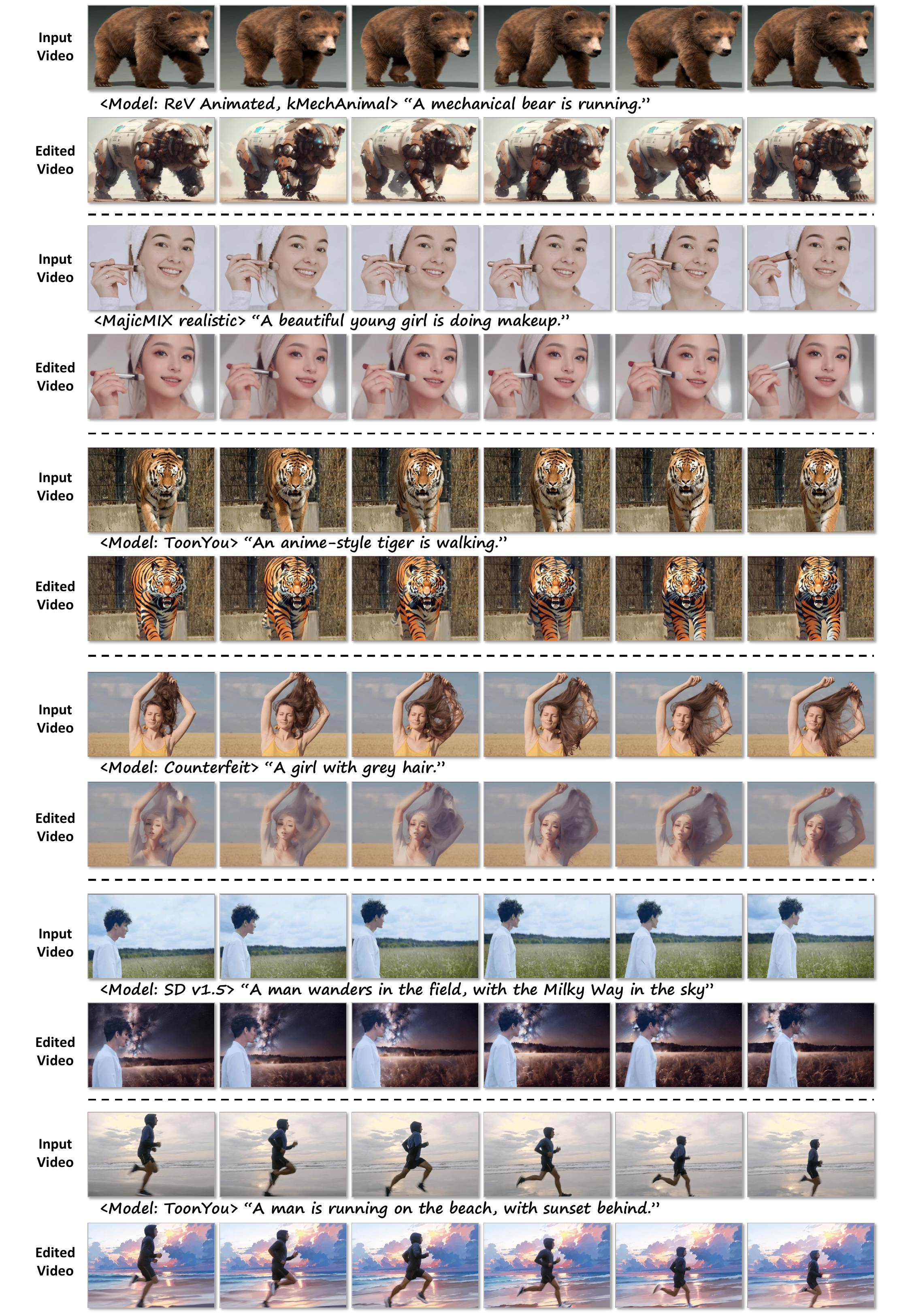}}
    \caption{\textbf{Visualized results of CCEdit.} $\langle \cdot \rangle$ indicates the personalized T2I model we used.
    }\label{fig:FigureSuppMoreVisualizations}
\end{figure*}

Fig.~\ref{fig:FigureSuppMoreVisualizations} shows several visualized results of CCEdit.




\subsection{Comprehensive Comparison}
\subsubsection{Compared Methods}
We compared our methods with eight state-of-the-art generative video editing methods: Tune-A-Video~\cite{wu2023tune}, vid2vid-zero~\cite{wang2023zero}, Text2Video-zero~\cite{khachatryan2023text2video}, FateZero~\cite{qi2023fatezero}, Pix2Video~\cite{ceylan2023pix2video}, ControlVideo~\cite{zhang2023controlvideo}, Rerender A Video~\cite{yang2023rerender}, and TokenFlow~\cite{geyer2023tokenflow}. The brief descriptions of these methods are as follows:

\textit{Tune-A-Video}~\cite{wu2023tune} propose the sparse attention mechanism to maintain the temporal coherence and optimize the network parameters through training on the source video. DDIM inversion~\cite{song2020denoising} is utilized to preserve the structure of input video.

\textit{Vid2vid-zero}~\cite{wang2023zero} utilizes off-the-shelf image diffusion models and employs the null-text inversion module~\cite{mokady2023null} for text-to-video alignment. Additionally, it incorporates a cross-frame modeling module to ensure temporal consistency and a spatial regularization module to maintain fidelity to the original video.

\textit{Text2Video-zero}~\cite{khachatryan2023text2video} introduces a method to enhance the latent codes of generated frames with motion dynamics, ensuring global scene and temporal consistency in the background. Additionally, it reprograms frame-level self-attention through cross-frame attention, focusing each frame on the first one to maintain the context, appearance, and identity of the foreground object.

\textit{FateZero}~\cite{qi2023fatezero} proposes to capture intermediate attention maps during inversion process, enhancing structural and motion information retention, and employs a novel spatial-temporal attention mechanism in the denoising UNet for improved frame consistency. 

\textit{Pix2Video}~\cite{ceylan2023pix2video} involves two steps to conduct generative video editing: initially, using a structure-guided (e.g., depth) image diffusion model to edit an anchor frame based on text prompts, followed by a key step of progressively propagating these edits to subsequent frames. This is done via self-attention feature injection, adapting the core denoising phase of the diffusion model. Adjustments are then made to the latent code of each frame before continuing the process.

\textit{ControlVideo}~\cite{zhang2023controlvideo} leverages ControlNet~\cite{zhang2023adding} to ensure the structural consistency from input video clips. In addition, it introduces full cross-frame interaction in self-attention modules for appearance coherence, an interleaved-frame smoother to reduce flickering through frame interpolation.

\textit{Rerender A Video}~\cite{yang2023rerender} propose to tackle the task of video editing by two parts: key frame translation and full video translation. Initially, it employs an adapted diffusion model to generate key frames, applying hierarchical cross-frame constraints to ensure coherence in shapes, textures, and colors. Subsequently, the framework extends these key frames to other frames using temporal-aware patch matching and frame blending techniques.

\textit{TokenFlow}~\cite{geyer2023tokenflow} propose the idea that the edited features convey the same inter-frame correspondences and redundancy as the original video features. Therefore, it propagates diffusion features based on inter-frame correspondences inherent in the model to ensure consistency in the diffusion feature space.

During the evaluation, all the videos consist of 17 frames at 6fps. We select depth maps as the structural representation. Additionally, to ensure fairness, the base model for all methods is Stable Diffusion v1.5.

\subsubsection{Qualitative Results}
\begin{figure*}[t]
\centerline{\includegraphics[width=0.9\linewidth]{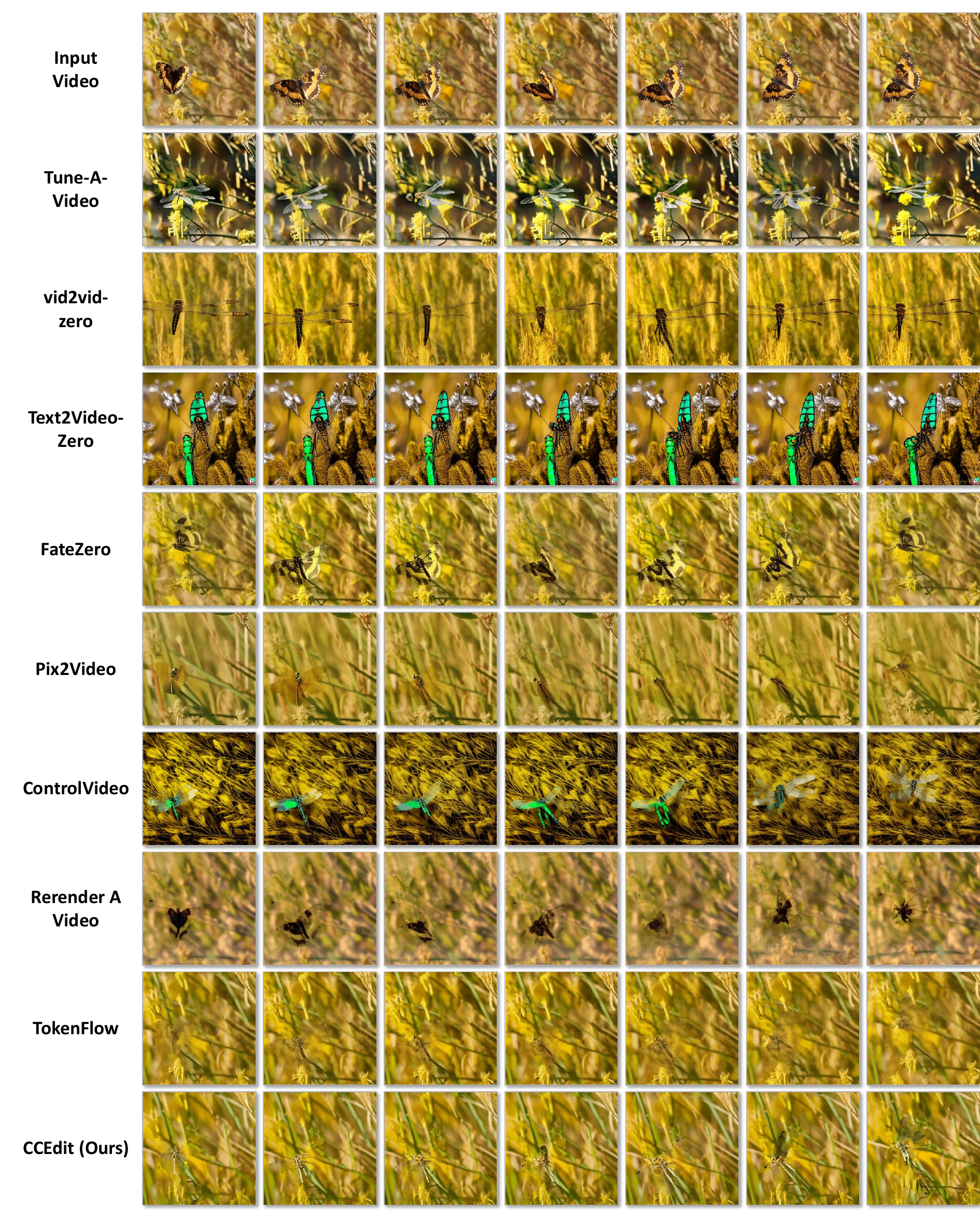}}
    \caption{\textbf{Qualitative comparison of different methods.} The target prompt is ``A dragonfly with shimmering wings perches on a plant amidst a field of golden grass.''
    }\label{fig:FigureSuppComparisonButterfly}
\end{figure*}

\begin{figure*}[t]
\centerline{\includegraphics[width=0.9\linewidth]{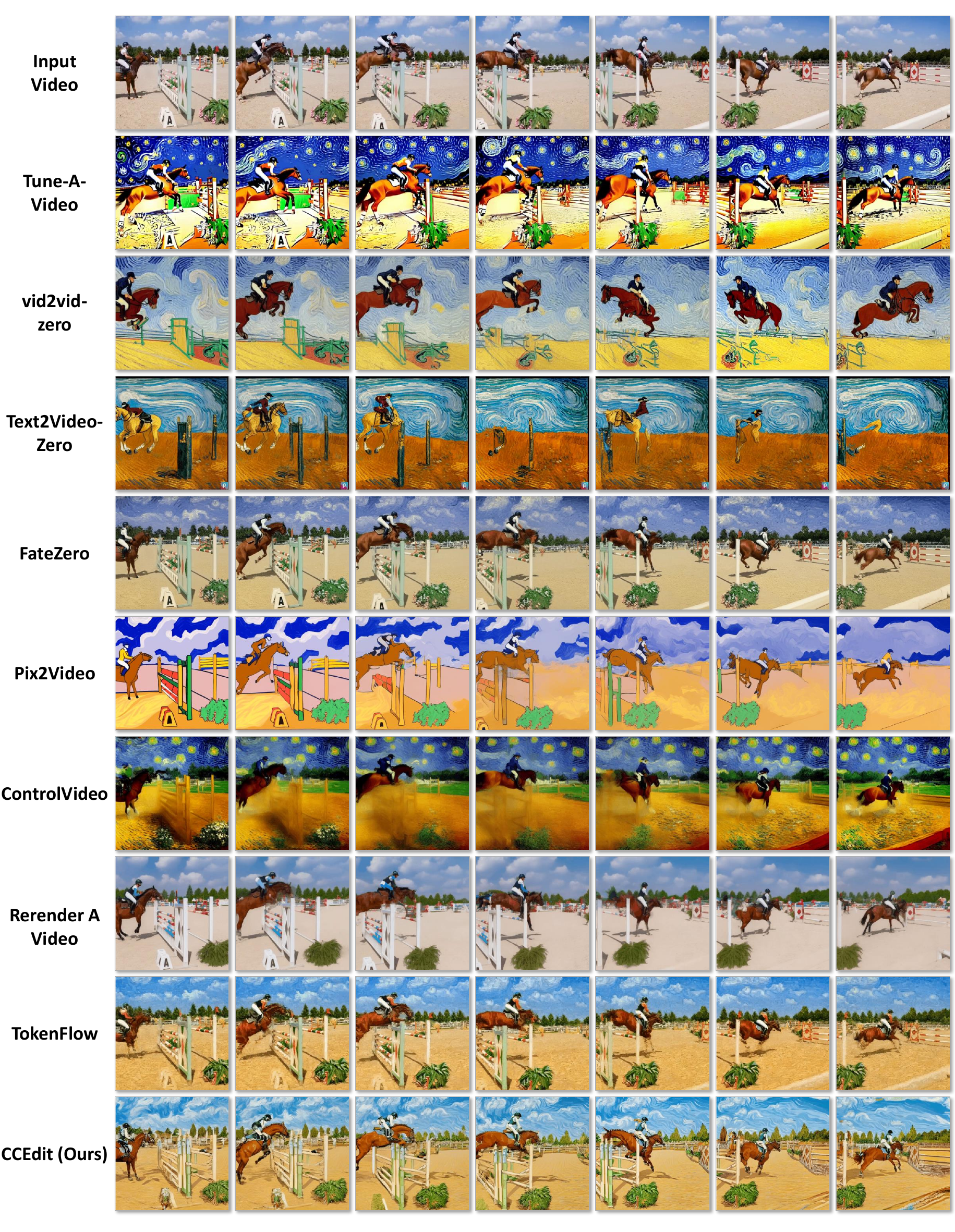}}
    \caption{\textbf{Qualitative comparison of different methods.} The target prompt is ``A rider on a horse jumping over an obstacle in an equestrian competition, rendered in Van Gogh style with swirling skies and vibrant colors.''
    }\label{fig:FigureSuppComparisonHorsejump}
\end{figure*}

The qualitative results for two videos are presented in Fig.~\ref{fig:FigureSuppComparisonButterfly} and Fig.~\ref{fig:FigureSuppComparisonHorsejump}. 
It can be observed that Tune-A-Video achieves effective editing that aligns well with the specified prompts, but falls short in maintaining temporal consistency and tends to produce overly contrasted images, possibly due to overfitting to the source video and excessively high default classifier-free guidance settings. Vid2vid-zero, Text2Video-Zero, and Pix2Video also struggle with insufficient temporal coherence. While FateZero exhibits better temporal coherence, its editing accuracy is not optimal. ControlVideo, despite its reasonable editing accuracy and temporal coherence, lacks a natural feel in its edited videos due to its global attention mechanism and interleaved-frame smoother technique. 
Rerender A Video demonstrates a limitation in executing precise edits, potentially due to an excessive dependence on detailed structural control mechanisms (line drawing and Canny edge of ControlNet). 
Such mechanisms restrict the method predominantly to minor stylistic alterations.
TokenFlow achieves stable results in both temporal coherence and editing accuracy, yet it still encounters blurring issues in scenes with significant object motion or rapid camera movements (see the horse legs in Fig.~\ref{fig:FigureSuppComparisonHorsejump}). 
At last, our approach demonstrates a notable capacity for sustaining temporal consistency, coupled with achieving exceptional accuracy in editing.

\subsubsection{Quantitative Results}
\begin{figure*}[t]
\centerline{\includegraphics[width=0.9\linewidth]{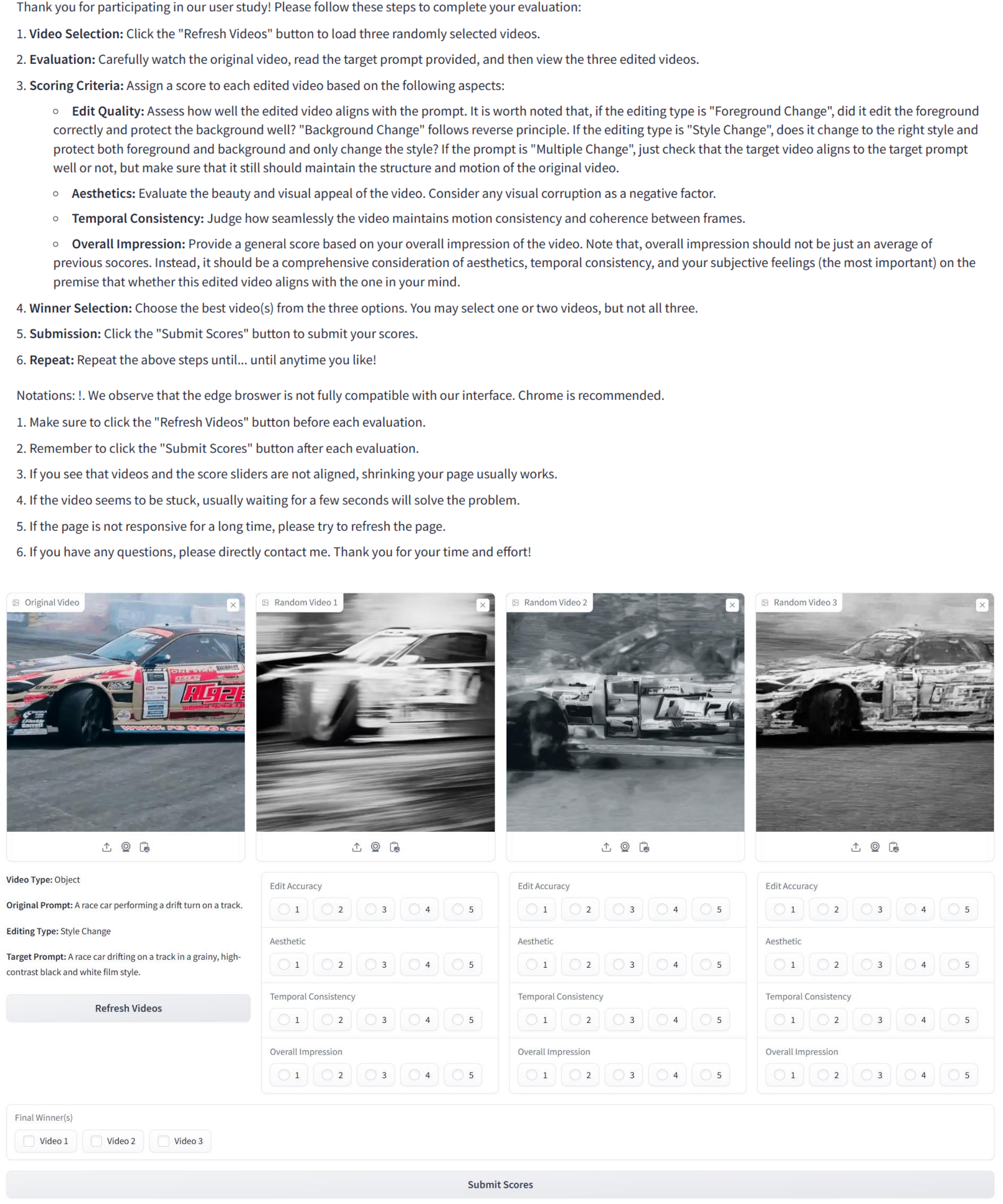}}
    \caption{\textbf{Illustration of the interface to conduct user study.} Initially, we provided a description of the evaluation criteria at the top of the page, along with corresponding notes for consideration. Additionally, to reduce user burden and avoid the confusion of displaying multiple videos simultaneously, for each video's editing result, we randomly selected three from all nine options (eight comparative methods and our method) for users to rate on various criteria (Edit Accuracy, Aesthetic, Temporal Consistency, and Overall Impression). Finally, users were asked to choose the winner(s) among the three videos. Selecting multiple winners was allowed (up to two), but choosing none or all three was not permitted. Zoom in to see details. 
    }\label{fig:FigureSuppUserstudyInterface}
\end{figure*}

\myparagraph{Automatic Metrics.} 
Our evaluation metrics include two aspects of both \textit{automatic ones} and \textit{user study} results. 
Automatic metrics are mainly conducted through the trained CLIP~\cite{radford2021learning,hessel2021clipscore,kirstain2023pick} model, similar to previous methods~\cite{wu2023tune,qi2023fatezero,ceylan2023pix2video,zhang2023controlvideo}. 
Specifically, \textit{``Tem-Con''} evaluates the temporal consistency of edited frames by calculating the similarity between successive frame pairs. Meanwhile, \textit{``Tex-Ali''} quantifies frame-wise editing accuracy, represented as the cosine similarity between edited frames and target prompts.
Additionally, the \textit{PickScore}~\cite{kirstain2023pick} is incorporated to predict the aesthetic quality and user preference of the edited videos.
Regarding the user study, we designed an interface and invited 33 volunteers to score the videos and pickup the winners, receiving a total of 1119 ratings. Each rating corresponds to various aspects of a single video. Specifically, the aspects to be rated include: \textit{``Editing Accuracy''}, representing whether the edited video accurately achieves the intended meaning of the target prompt;  \textit{``Aesthetics''}, denoting the visual appeal of the edited video;  \textit{``Temporal Consistency''}, indicating whether the video maintains coherence over time; and \textit{``Overall Impression''}, which reflects the subjective overall rating of the video. The interface is illustrated in Fig.~\ref{fig:FigureSuppUserstudyInterface}.

\myparagraph{Results of Automatic Metrics.} The results are illustrated in Tab.~\ref{table:Supp_Quantitative_AutomaticMetrics}. 
Although our method ranked second in temporal consistency and first in text alignment in the table of user study presented in the main text, it did not particularly stand out in terms of corresponding objective metrics. This observation has been noted in many previous works~\cite{zhang2023towards,molad2023dreamix,wu2023cvpr}, further emphasizing the significance of more advanced objective automatic metrics for the development of this field. Finally, our method achieved the best performance in the CLIP-based scoring function, PickScore, an indicator of human preference, demonstrating its superior alignment with human subjective perceptions.

\begin{table}[t]
\footnotesize
  \centering
  \begin{tabular}{l@{\hspace{7pt}}c@{\hspace{7pt}}c@{\hspace{7pt}}c}
    \toprule
    
    Method     & Tem-Con $\uparrow$  & Tex-Ali $\uparrow$ & Pick $\uparrow$ \\
    
    \midrule
    
    Tune-A-Video~\cite{wu2023tune}         & 0.937 & 0.284 & 0.206  \\
    vid2vid-zero~\cite{wang2023zero}        & 0.933 & 0.284 & 0.209  \\
    Text2Video-Zero~\cite{khachatryan2023text2video}           & 0.949 & 0.262 & 0.203 \\
    FateZero~\cite{qi2023fatezero}            & 0.942 & 0.245 & 0.205 \\
    Pix2Video~\cite{ceylan2023pix2video}           & 0.939 & \textbf{0.285} & 0.208 \\
    ControlVideo~\cite{zhang2023controlvideo}        & \textbf{0.950} & \textbf{0.285} & 0.210  \\
    Rerender A Video~\cite{yang2023rerender}    & 0.928 & 0.247 & 0.201 \\
    TokenFlow~\cite{geyer2023tokenflow}           & 0.949 & 0.270 & 0.210 \\
 
    \midrule
    
    CCEdit (Ours)         & 0.936 & 0.281 & \textbf{0.213} \\

    \bottomrule
  \end{tabular}  
  \caption{\textbf{State-of-the-art comparison of automatic metrics.}
  ``Tem-Con'' represents temporal consistency, ``Text-Ali'' indicates textural alignment, and ``Pick'' represents to the PickScore~\cite{kirstain2023pick}.
  }
   \label{table:Supp_Quantitative_AutomaticMetrics}
\end{table}

\begin{table}[t]
\footnotesize
  \centering
  \begin{tabular}{lccc}
    \toprule
    
    Method     & Pre-Processing   & Inference & Total \\
    
    \midrule
    
    Tune-A-Video~\cite{wu2023tune}                      & 545 & 22 & 567 \\
    vid2vid-zero~\cite{wang2023zero}                    & 148 & 230 & 378 \\
    Text2Video-Zero~\cite{khachatryan2023text2video}    & 0 & 28 & 28 \\
    FateZero~\cite{qi2023fatezero}                      & 199 & 42 & 241 \\
    Pix2Video~\cite{ceylan2023pix2video}                & 0 & 188 & 188 \\
    ControlVideo~\cite{zhang2023controlvideo}           & 0 & 56 & 56 \\
    Rerender A Video~\cite{yang2023rerender}            & 76 & 96 & 172 \\
    TokenFlow~\cite{geyer2023tokenflow}                 & 182 & 27 & 209 \\
 
    \midrule
    
    CCEdit (Ours)                                       & 134 & 46 & 170 \\

    \bottomrule
  \end{tabular}  
  \caption{\textbf{Runtime comparison (seconds).}}
   \label{table:Supp_Runtime}
\end{table}

\subsubsection{Runtime Analysis}
Tab.~\ref{table:Supp_Runtime} presents the runtime of various methods, detailing the time spent on pre-processing, inference, and the total duration, respectively. 
Pre-processing includes tasks of fine-tuning on the source video, performing inversion operations, caching attention maps, key frame editing, and others. The inference time represents the duration of the sampling process, along with all the associated operations.
Overall, the time consumed by our method is not lengthy compared to other video editing techniques.
It is worth noted that in our method, the time spent on key frame editing using PnP~\cite{tumanyan2023plug}) during pre-processing constitutes the majority of the total time, while the actual sampling time is relatively brief. 
It's attributed to the absence of any inversion and attention map operations. The only additional computational overhead arises from the extra network parameters introduced during the network forward process. 
In practical applications, one can opt for more advanced and lightweight image editing methods or manually make fine adjustments, thereby achieving the desired trade-off.
This further demonstrates the practicality and flexibility of our approach.




\begin{figure}[t]
\centerline{\includegraphics[width=0.83\linewidth]{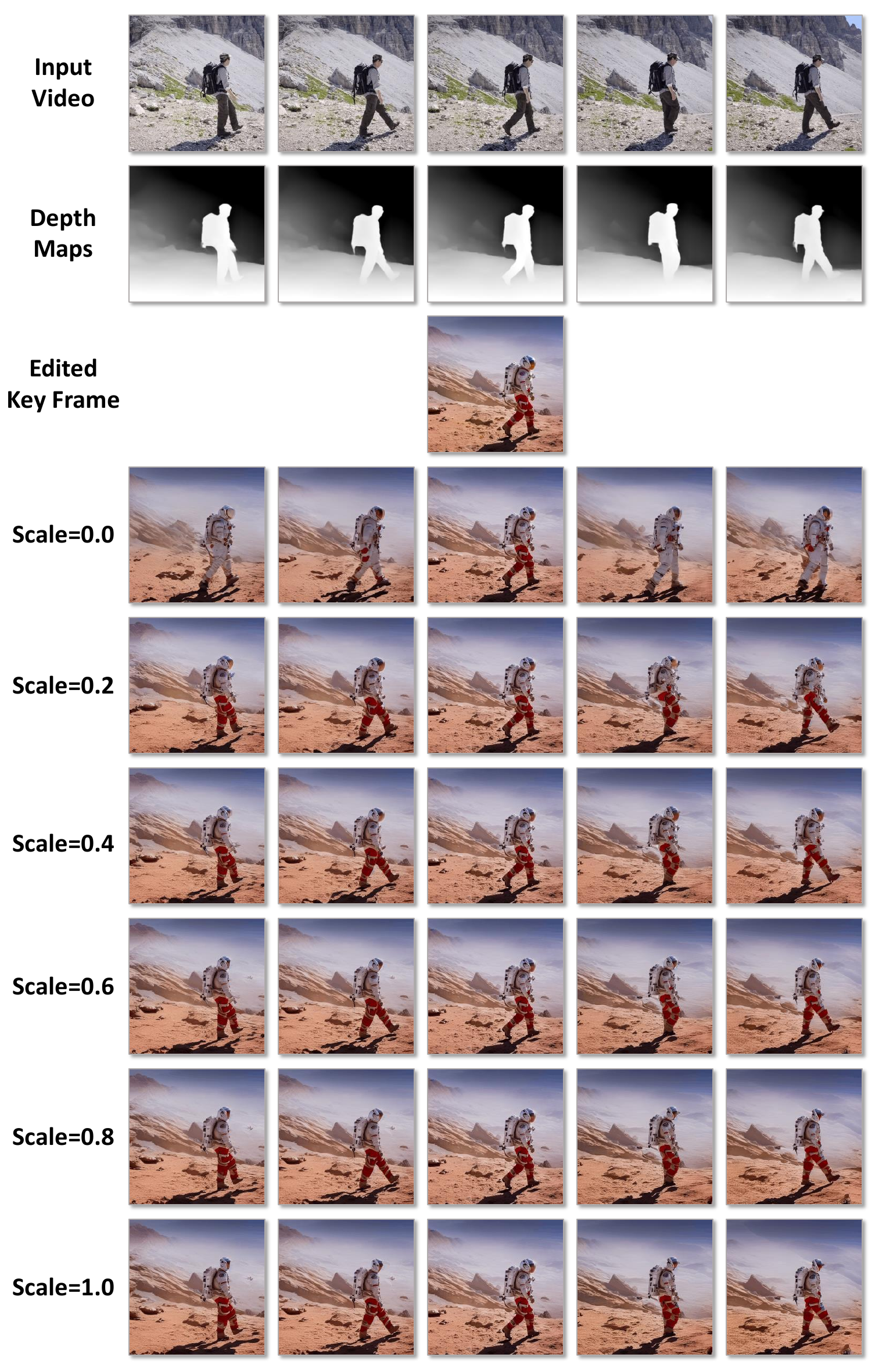}}
    \caption{\textbf{Results at different scales of structure branch.} The target prompt is ``An astronaut with a jetpack floats above a Martian landscape, with red rocky terrains and tall, alien-like mountains in the backdrop.''
    }\label{fig:FigureSuppAblationControlScaleStructure}
\end{figure}
\begin{figure}[t]
\centerline{\includegraphics[width=0.83\linewidth]{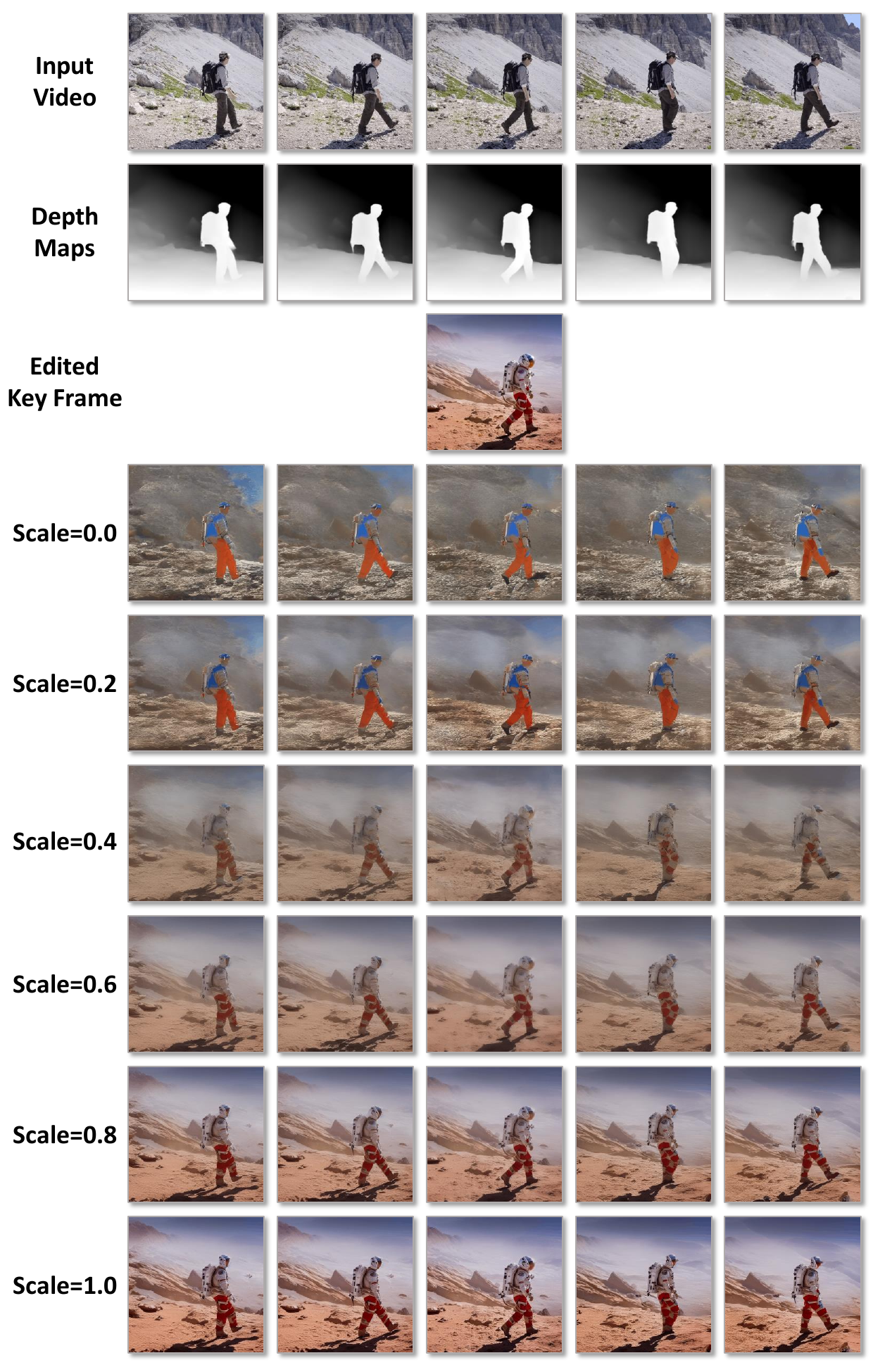}}
    \caption{\textbf{Results at different scales of appearance branch.} The target prompt is ``An astronaut with a jetpack floats above a Martian landscape, with red rocky terrains and tall, alien-like mountains in the backdrop.''
    }\label{fig:FigureSuppAblationControlScaleAppearance}
\end{figure}

\subsection{Study on Control Scales}
\myparagraph{Structure Branch.}
Sometimes, the appearance of the edited key frame may structurally differ from the corresponding structure representation of the original video. 
Since the features of the structure branch are injected into the main branch through summation, the intensity of structure information infusion can be adjusted by modifying the coefficients (named control scale) applied to the features during this summation process.
In such cases, reducing the control scale of the structure branch could help. This adjustment lessens its structural constraints on the results, allowing for a greater reliance on the information provided by the appearance branch and adherence to the coherence adjustments made by the temporal layers. The visualized results are shown in Fig.~\ref{fig:FigureSuppAblationControlScaleStructure}. 
It can be observed that in the edited key frame, the astronaut's silhouette appears markedly larger than that of the original person, a consequence of the voluminous spacesuit.
When the structure control is relatively high (0.6$\sim$1.0), the editing results show that the center frame remain consistent with the edited frame, while the structure of other frames is overly constrained by the structure representation. At a control scale of 0, the loss of structure information leads to the astronaut being unable to move correctly. However, with a moderate control scale (0.2$\sim$0.4), a better trade-off is achieved in terms of appearance, structure, and motion. Note that in comparisons with other methods, to ensure fairness, our method consistently employed a control scale of 1.

\myparagraph{Appearance Branch.}
Since the features of the Appearance Branch are also injected into the main branch through summation, the intensity of appearance information infusion can similarly be adjusted by tuning the summation coefficients of the appearance branch. The results are shown in Fig.~\ref{fig:FigureSuppAblationControlScaleAppearance}. At a lower control scale (0$\sim$0.2), the influence of appearance information is minimal, barely impacting the edited video. When the control scale is moderate (0.4$\sim$0.6), appearance information begins to play a role. However, possibly due to conflicts with the priors of the main branch, this results in a somewhat dull and dark color tone in the visuals. Conversely, at a higher control scale (0.8$\sim$1.0), appearance information exerts a decisive control over the overall appearance of the edited video.




\subsection{Study on Text Prompt}
\begin{figure}[t]
\centerline{\includegraphics[width=0.9\linewidth]{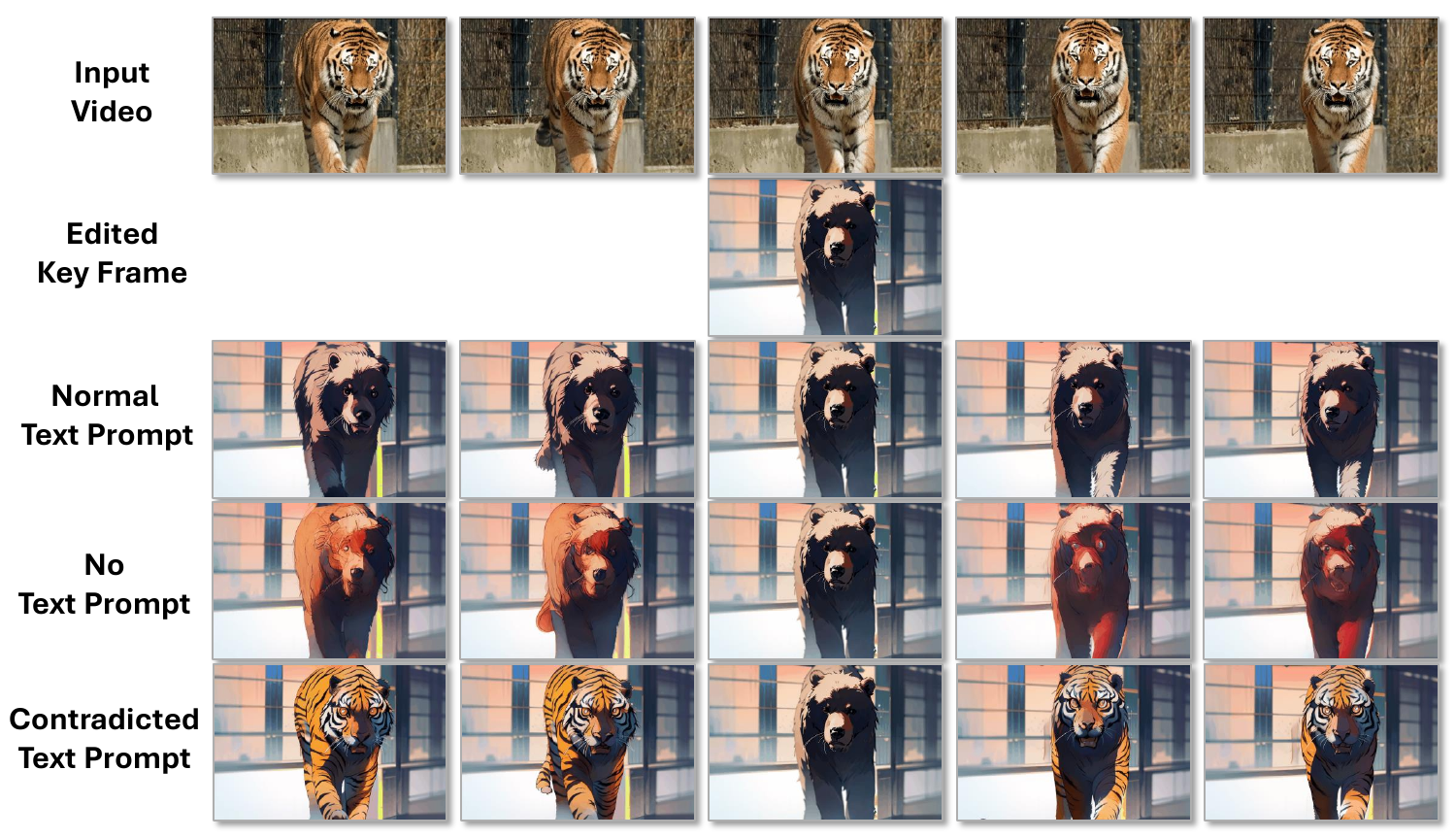}}
    \caption{\textbf{Illustration of results with different text prompts}. The normal prompt is ``A bear is walking''. The contradicted text prompt is ``A tiger is walking''. The ``ToonYou'' personalized T2I model is used.}
    \label{fig:FigureSuppAblationTextPrompts}
\end{figure}

Another point worth exploring is whether text prompts are still necessary after introducing appearance control. To address this, we conducted a visual experiment. As shown in Fig.~\ref{fig:FigureSuppAblationTextPrompts}, providing a normal text prompt leads to correct results, whereas the absence of any text prompt results in significant distortions in the generated output. When given a text prompt that contradicts the appearance information, only the center frame retains the appearance information, while the other frames are controlled by the text prompt. 
Consequently, the conclusion is that text prompts are still necessary within this framework. We believe this may be due to the weights of the main branch and the structure branch being frozen during the training process. As a result, the entire editing process seems to involve the appearance branch exerting more detailed control over the image after the text prompt has already provided a coarse guide.

\section{Limitation and Future Works}
\subsection{Structural Deviation}
As described in the main text, a primary challenge that needs addressing in our video editing approach is the structural deviation (also the major issue mentioned in TokenFlow~\cite{geyer2023tokenflow}) between the input and target videos. This deviation could stem from semantic changes inherent to the target or from alterations in the target's behavior. For instance, transitioning from a ``cute rabbit'' to a ``fierce tiger'' is challenging due to their fundamentally different physiological structures. Most existing methods struggle to overcome this hurdle and often only manage to modify their textural appearance. In our approach, adjusting the scale coefficient of structure branch and employing coarser-grained structure representations (like the skeleton) may alleviate this issue to some extent, but we believe it doesn't fundamentally solve the problem. 
Achieving changes in the target's behavior, such as transforming a ``running bear'' into a ``dancing bear'', is even more challenging. This complexity arises primarily because most contemporary generative video editing methods employ Text-to-Image (T2I) models at the image level. These models, devoid of prior knowledge concerning actions, encounter difficulties in editing motion.

We posit that a promising approach could be to integrate a pre-trained T2V (text-to-video) model, cleverly utilizing its priors to tackle these challenges.



\subsection{Heavy Appearance and Structure Branch}
In CCEdit, the appearance and structure branch utilize two heavy encoder to extract features, consisting significant amount of parameters.
This may be unnecessary and could lead to issues such as increased GPU memory consumption and longer editing times during use. In the future, we plan to explore the adoption of more lightweight networks~\cite{mou2023t2i,controlnetxs} to address these concerns.

\subsection{Flickering Problem.} 
We observed flickering in videos with higher frame rates or after frame interpolation, especially noticeable in high-frequency fine texture details. 
This is primarily attributed to our video editing operations being performed in the latent domain encoded by the 2D autoencoder.
Introducing additional temporal layers in the autoencoder~\cite{blattmann2023align} is an promising way to solve this problem.

\clearpage
\clearpage

